\documentclass[review,5p]{elsarticle}
\usepackage{graphicx}

\usepackage{caption}
\usepackage{subcaption}
\usepackage{color}
\usepackage{tikz}
\usetikzlibrary{arrows}
\usetikzlibrary{calc}
\usepackage{float}
\usepackage{hyperref}

\usepackage{array}
\usepackage{multirow}
\usepackage{arydshln}

\usepackage{tabularx}

\usepackage{amsmath}
\usepackage{amssymb}
\usepackage{stfloats}
\usepackage{nicefrac}

\usepackage[acronym]{glossaries}

\usepackage{algorithm}
\usepackage{algpseudocode}

\usepackage{caption}

\usepackage{amsmath}
\usepackage{mathtools}

\usepackage{enumitem}
\usepackage{array}

\journal{}

\begin{document}

\begin{frontmatter}

\title{Fast and Accurate Light Field Saliency Detection through Deep Encoding}

\author{Sahan Hemachandra$^*$, Ranga Rodrigo$^*$, Chamira U. S. Edussooriya$^{*,\#}$}
\address{$^*$Department of Electronic and Telecommunication Engineering, University of Moratuwa, Sri Lanka \\
$^{\#}$Department of Electrical and Computer Engineering, Florida International University, Miami, FL, USA}

\begin{abstract}
\noindent Light field saliency detection---important due to utility in many vision tasks---still lacks speed and can improve in accuracy. Due to the formulation of the saliency detection problem in light fields as a segmentation task or a memorizing task, existing approaches consume unnecessarily large amounts of computational resources for training, and have longer execution times for testing. We solve this by aggressively reducing the large light field images to a much smaller three-channel feature map appropriate for saliency detection using an RGB image saliency detector with attention mechanisms. We achieve this by introducing a novel convolutional neural network based features extraction and encoding module. Our saliency detector takes $0.4$ s to process a light field of size $9\times9\times512\times375$ in a CPU and is significantly faster than state-of-the-art light field saliency detectors, with better or comparable accuracy. Furthermore, model size of our architecture is significantly lower compared to state-of-the-art light field saliency detectors. Our work shows that extracting features from light fields through aggressive size reduction and the attention mechanism results in a faster and accurate light field saliency detector leading to near real-time light field processing.
\end{abstract}


\end{frontmatter}

\section{Introduction}
\label{sec:intro}
Light fields capture both spatial and angular information of light emanating from a scene compared to spatial-only information captured by images. The additional angular information available with light fields paves the way for novel applications such as post-capture refocusing~\cite{Ng2005a,Dan2015,Jay2021} and depth-based filtering~\cite{Dan2007,Edu2015a,Liy2020}, which are not possible with images. Furthermore, light fields support numerous computer vision tasks which are traditionally based on images and videos~\cite{Lev1996,Dan2011b,Dong2013,Wu2017,Yu2017,Zel2017,Lu2018,Dan2019}.  

Saliency detection is a prerequisite for many computer vision tasks such as semantic segmentation, image retrieval, and scene classification. Saliency detection using light fields provides better accuracy compared to what is provided by RGB images, in particular, for challenging scenes having similar foreground and background, and complex occlusions~\cite{li2017saliency,zhang2015saliency}. However, data available with light fields (i.e., pixels per light field) are significantly higher than data available with a single RGB image, e.g., a light field having $9\times9$ sub-aperture images contains $81$ times more data (with the same resolution). Therefore, computational time of light field saliency detection algorithms is substantially higher compared to that of RGB image saliency detection algorithms~\cite{zhang2015saliency}. 

We can categorize existing light field saliency detectors in to three classes depending on the input: focal stack and all-focus image, RGB-D images, and light fields. In the first class, a set of two-dimensional (2-D) images focused at different depths, called a focal stack, and a sub-aperture image, called all-focus image, are used as the input. Here, a focal stack is generated from a light field using a refocusing algorithm~\cite{Ng2005a,Dan2015}, and this step acts as a preprocessing step with additional computations. Furthermore, focal stack generation requires human intervention because the number of 2-D images in a focal stack depends on a light field.  The second class employs RGB-D images consisting of an all-focus image and a depth map. In this case, the depth map is generated using a depth estimation algorithm~\cite{Tao2013,Wang2015,Chen2018} and incur additional computations. Similar to the generation of a focal stack, generation of a depth map is also a preprocessing step. Compared to these two classes, the third class employs a light field as the input without any preprocessing steps. Recent algorithms of these three categories predominately use convolutional neural networks (CNNs) to learn the relationship between the image features and saliency of light fields. Even though the available light field datasets are limited in size, we can freely augment focal stack and RGB-D data in the first two classes. On the other hand, inability to freely augment light field images prevents training deep CNNs from scratch in the third class. These constraints demand the use of pre-trained networks, of course, followed by fine tuning.

In this paper, we propose a novel feature \emph{extraction and encoding} (FEE) module for \emph{fast} light field saliency detection by employing an 2-D RGB image saliency detection algorithm. Our FEE module takes light field as the input (so, belongs to the third class), and provides an RGB encoded feature map. The proposed FEE module comprises of a CNN with five convolutional layers. We employ the 2-D saliency detector proposed in~\cite{zhao2019pyramid} with our FEE module. Furthermore, we employ the LYTRO ILLUM saliency detection dataset~\cite{zhang2020lightfield} and DUTLF-v2 dataset~\cite{DUTLF} for the training and testing the performance of our light field saliency detector. Experimental results obtained with five-fold cross validation confirms that our  saliency detector provides a \emph{significant improvement} in computational time with accuracy comparable or better  than state-of-the-art light field saliency detectors~\cite{zhang2020lightfield,zhang2020multi}. Furthermore, model size of our architecture is significantly lower compared to state-of-the-art light field saliency detectors leading to a lower memory requirement.

The paper is organized as follows. In Section~\ref{sec:relatedworks}, we review different approaches employed and algorithms proposed for saliency detection of RGB images and light fields. We present our light field saliency detector in detail in Section~\ref{sec:methodology}. In Section~\ref{sec:results}, we present experimental results to verify the accuracy and the speed of the proposed light field saliency detector. Finally, in Section~\ref{sec:cnfw}, we present conclusion and future works. 

\section{Related Works}
\label{sec:relatedworks}

\begin{figure*}[tbh!]
     \centering
     \begin{subfigure}[b]{0.70\linewidth}
         \centering
         \input{LF}
         \caption{Sub-aperture image array and a single view}
         \label{fi:LFa}
     \end{subfigure}%
     \hspace{2mm}
     \begin{subfigure}[b]{0.27\linewidth}
         \centering
         \input{LRSA}
         \caption{Lens array image}
         \label{fi:LFa}
     \end{subfigure}%
     
     \begin{subfigure}[b]{0.3\linewidth}
         \centering
         \includegraphics[width=0.95\linewidth]{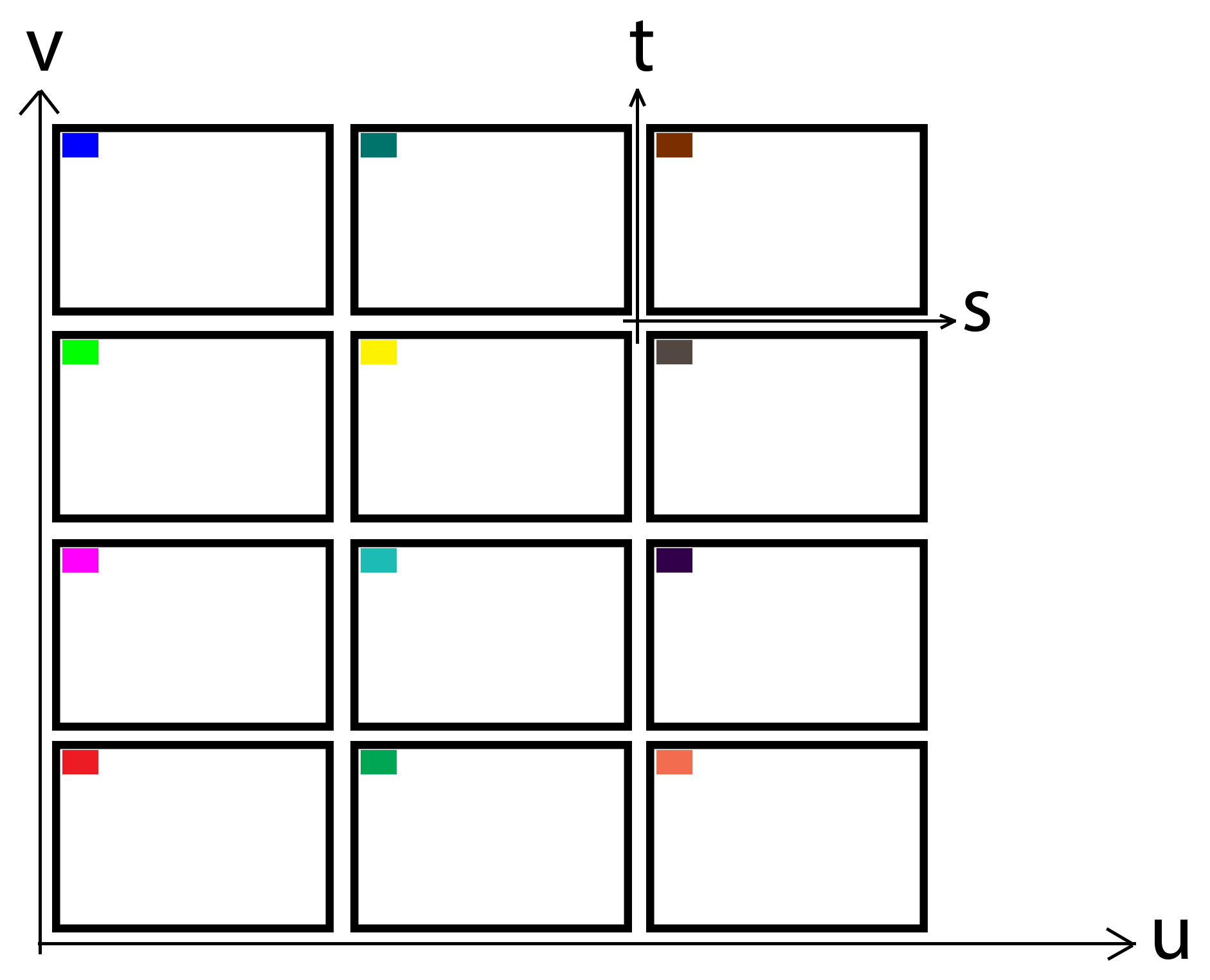}
         \caption{pixels in sub-aperture image array}
         \label{fi:LFb}
     \end{subfigure}%
     \hspace{5mm}
     \begin{subfigure}[b]{0.3\linewidth}
         \centering
         \includegraphics[width=0.95\linewidth]{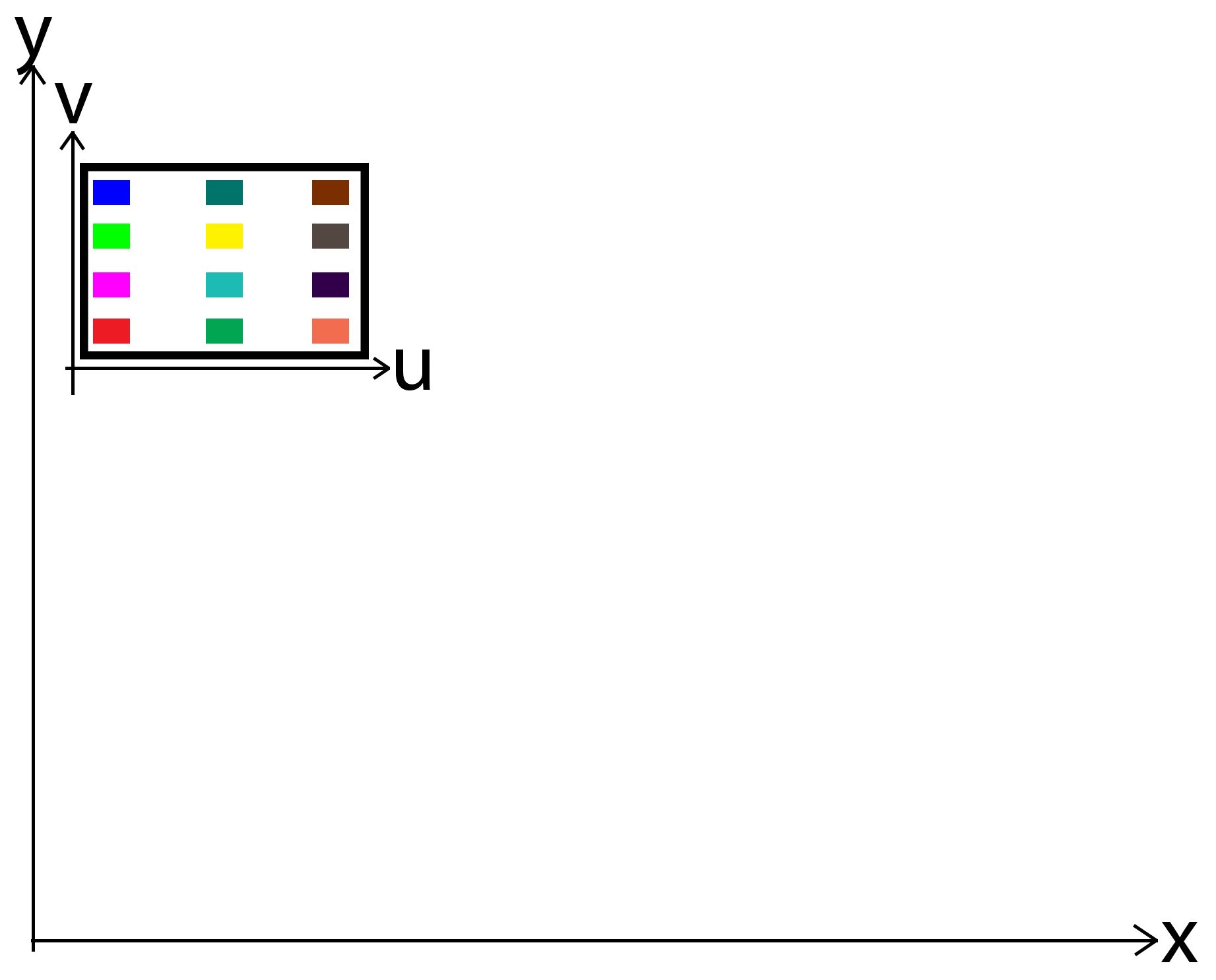}
         \caption{pixels in lens array image}
         \label{fi:LFc}
     \end{subfigure}
     \caption{A light field consists of $U\times V$ sub-aperture images as shown in (a), where each sub-aperture image has $S\times T$ pixels as shown in (b); (c) depicts how the pixels in sub-aperture images are arranged in order to obtain a micro lens array image shown in (d), where the first block of size $U\times V$ contains all the pixels of spatial position $(1,1)$ in each sub-aperture image.}
     
     \label{fi:LF}
\end{figure*}

\subsection{Saliency Detection on RGB images}
Saliency detection is one of the oldest problems in computer vision research and there have been many research done on various approaches for this task in the recent time. Earliest research~\cite{yang2013graph,cheng2015global,perazzi2012sal,zhu2014sal} were mainly based on handcrafted features like boundaries or contrast of the images to detect the most salient objects in the RGB images. \cite{yang2013graph} proposed a graph based manifold ranking algorithm for salience detection based on background and foreground cues. \cite{cheng2015global} proposed an regional contrast based algorithm, where global contrast and spatial weighted coherence scores are used simultaneously to accurately detect the salient objects. \cite{perazzi2012sal} introduced a contrast based approach using high dimensional Gaussian filters to unifying detect salience and complete contrast while \cite{zhu2014sal} used background priors to detect the salient regions on images. Even though these methods are less computationally expensive, they tends to fail in complex backgrounds. 

With the popularity of deep learning in the last decade, many approaches based on deep learning has been introduced for the RGB saliency detection using neural networks. There is rich body of work in  saliency detection in RGB images: pyramidal, feature based, recurrent network based, and attention based. Most non-recurrent methods use VGG-16- or VGG-19-like feature extractors \cite{simonyan2014very} pre-trained on ImageNet dataset  for feature extraction. Pyramidal saliency detectors \cite{zhang2018progressive, li2016deep, zhao2019pyramid} have the advantage of the ability to use information from multiple layers. Some that build up on CNN feature computers defer the actual saliency detection to latter layers or combine features from many layers \cite{zhao2015saliency, liu2018picanet}. Methods that employ recurrent networks generally work well \cite{wang2016saliency, kuen2016recurrent} with the possible disadvantage of slowness. RGB saliency detectors greatly benefit from attention models, by focusing on features that truly capture saliency without the interference of unnecessary features.
Although these methods show success in RGB images, they are unsuitable for direct use with light field images because their architecture and input are not specifically designed to extract the geometry information of light fields embedded in angular dimensions. This information  is vital to improve the quality of predicted saliency maps.

\subsection{Saliency Detection on Light Fields}
\label{ssec:existing_methods}
Light field saliency detection~\cite{li2017saliency} improves the accuracy of saliency detection in challenging scenes having similar foreground/background and complex occlusions. This improvement achieves in~\cite{li2017saliency} by exploiting the refocusing capability available with light fields which provides focusness, depths, and objectness cues. \cite{zhang2015saliency} employs depth map, all focus image and focal stack available with a light field for saliency detection. \cite{zhang2017saliency} further exploits light field flow fields over focal slices, and multi-view sub-aperture images improve the accuracy in saliency detection by enhancing depth contrast. \cite{li2015weighted} employs a dictionary learning based method to combine various light field features for a universal saliency detection framework using sparse coding. This method handles various types of input data by building saliency and non-saliency dictionaries using focusness cues of focal stack as features for light fields. All these methods works on super-pixel level features of light fields, and do not exploit high-level semantic information properly in order to have robust performance in complex scenarios.

\vspace{-1ex}
\subsection{Light Field Saliency Detection with Deep Learning}
\label{ssec:dl_on_lightfields}

Recent advances in light field saliency detection successfully use deep neural networks. \cite{Wang_2019_ICCV} introduced a two-stream neural network architecture with two VGG-19 feature extractors and ConvLSTM-based attention module to process the all-focus image and focal stack to generate saliency maps. The saliency detection model in~\cite{piao2019deep} use a deep neural network pipeline containing light field synthesising network using center view and a light field driven saliency detection network to detect salient objects in single view images. Similarly, \cite{zhang2020multi} employed a multi-task collaborative network (MTCN) for light field saliency detection with two streams for central view image and multi-view images by exploring the spatial, depth and edge information in different parts of their neural network with the help of a complicated loss function with different components for different parts of the network. \cite{zhang2020lightfield} introduced a ``model angular changes block'' to process light field images with a modified version of Deeplabs v-2 segmentation network (LFNet), which is a computationally heavy backbone, considering the similarity between the segmentation and saliency detection. On the other hand, the suitability of a semantic segmentation network, not specifically trained on light fields,  may affect accuracy. \cite{DUTLF} introduced, two-stream network containing teacher and student network to detect salient objects, exploiting focal stack and all focus image in their respective streams of the network. Most of these methods have the inherent disadvantage of slowness due to use of heavy segmentation networks, several feature extractors, recurrent blocks and several streams. Furthermore, full light field data are employed for most of the parts and layers of the neural networks hindering the speed.

\section{Proposed Light-Field Saliency Detection Architecture}
\label{sec:methodology}
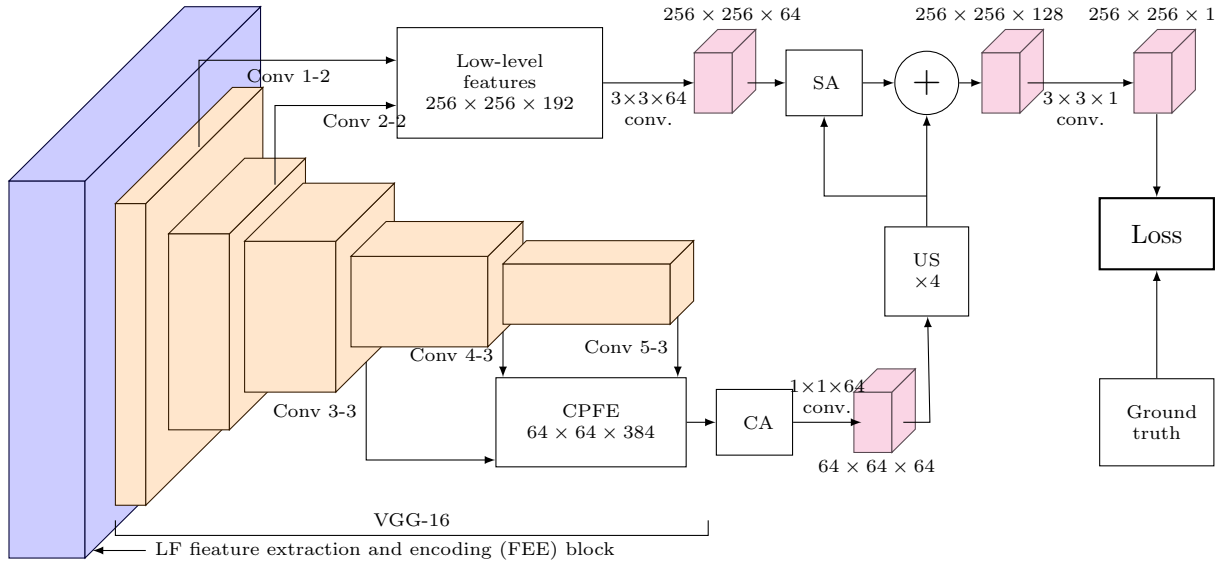
\begin{figure*}[t!]
    \begin{center}
        \begin{tikzpicture}

\setlength{\baselineskip}{5em}

\tikzstyle{bluebox}=[draw=blue!20!black,fill=blue!20]
\tikzstyle{orangebox}=[draw=orange!20!black,fill=orange!20]
\tikzstyle{magentabox}=[draw=magenta!20!black,fill=magenta!20]

\tikzstyle{labelnode}=[align=center, execute at begin node=\setlength{\baselineskip}{0.8em}]

\newcommand{\cube}[5][]
{
	\pgfmathsetmacro{\cubex}{(#2)}
	\pgfmathsetmacro{\cubey}{{#3}}
	\pgfmathsetmacro{\cubez}{{#4}}
	\draw[#5] (0,0,0) -- ++(-\cubex,0,0) -- ++(0,-\cubey,0) -- ++(\cubex,0,0) -- cycle;
	\draw[#5] (0,0,0) -- ++(0,0,-\cubez) -- ++(0,-\cubey,0) -- ++(0,0,\cubez) -- cycle;
	\draw[#5] (0,0,0) -- ++(-\cubex,0,0) -- ++(0,0,-\cubez) -- ++(\cubex,0,0) -- cycle;
}
\begin{scope}[xshift=0.4cm, yshift=0.2cm]
	\cube{1}{5}{6}{bluebox};
\end{scope}

\begin{scope}[xshift=1.2cm, yshift=-0.1cm]
	\cube{0.4}{4}{4}{orangebox};
	\node at (+0, -4.6) [anchor=west] {\scriptsize LF fieature extraction and encoding (FEE) block};
	\draw [-latex] (0, -4.6) -- ++(-0.7, 0);
\end{scope}
\begin{scope}[xshift=2.3cm, yshift=-0.5cm]
	\cube{0.8}{2.6}{2.6}{orangebox};
	\draw (-1.5, -3.8) -- ++(0,-0.2) -- ++ (7.8,0) node[midway, above] {\scriptsize VGG-16} -- ++(0, 0.2);
\end{scope}
\begin{scope}[xshift=3.7cm, yshift=-0.6cm]
	\cube{1.2}{2}{2}{orangebox};
\end{scope}
\begin{scope}[xshift=5.7cm, yshift=-0.8cm]
	\cube{1.8}{1.2}{1.2}{orangebox};
\end{scope}
\begin{scope}[xshift=8.1cm, yshift=-0.9cm]
	\cube{2.2}{0.8}{0.8}{orangebox};
\end{scope}

\node (cpfe) at (5.8, -3) [anchor=west, draw=black, text width=1.8cm, align=center, inner sep=10pt, labelnode] {\scriptsize CPFE $64\times 64 \times 384$};
\node (ca) at (8.7, -3) [anchor=west, draw=black, text width=0.3cm, align=center, inner sep=10pt] {\scriptsize CA};

\draw[-latex] (8.2, -1.6) coordinate (conv53) -- (cpfe.north -| conv53) node[midway, anchor=east] {\scriptsize Conv 5-3};
\draw[-latex] (5.9, -1.8) coordinate (conv43) -- (cpfe.north -| conv43) node[midway, anchor=east] {\scriptsize Conv 4-3};

\draw[-latex] (4.1, -2.2) |- ($(cpfe.west) + (0, -0.5)$) node[near start, anchor=east] {\scriptsize Conv 3-3};
\draw[-latex] (cpfe) -- (ca);

\begin{scope}[shift={($(ca.east)+(1.3cm,0.4cm)$)}]
	\cube{0.5}{0.8}{0.8}{magentabox};
	\node at (-0.2,-1) {\scriptsize $64\times 64 \times 64$};
\end{scope}

\draw[-latex] (ca.east) -- ++(0.9,0) node[midway, above, text width=1cm, align=center, labelnode] {\scriptsize $1\times 1\times 64$ conv.};
\node (upsamp) at ($(ca.east) + (1.2,2)$) [anchor=west, draw=black, text width=0.4cm, align=center, inner sep=10pt, labelnode] {\scriptsize US $\times4$};
\draw[-latex] (ca.east) ++(1.45,0)--++(.35,0) -- (upsamp);

\node (llf) at (4.5, 1.5) [anchor=west, draw=black, text width=2cm, align=center, inner sep=10pt, labelnode] {\scriptsize Low-level features $256\times 256 \times 192$};

\draw[-latex] (1.9, 0.65) |- ($(llf.west) + (0, 0.3)$) node[midway, xshift=0.5cm, yshift= -0.2cm, anchor=west] {\scriptsize Conv 1-2};
\draw[-latex] (2.9, 0.15) |- ($(llf.west) + (0, -0.3)$) node[midway, xshift=0.5cm,  yshift= -0.2cm, anchor=west] {\scriptsize Conv 2-2};

\begin{scope}[shift={($(llf.east)+(1.7cm,0.4cm)$)}]
	\cube{0.5}{0.8}{0.8}{magentabox};
	\node at (0,0.5) {\scriptsize $256\times 256 \times 64$};
\end{scope}
\draw[-latex] (llf.east) -- ++(1.2,0) node[midway, below, text width=1cm, align=center, labelnode] {\scriptsize $3\times 3\times 64$ conv.};
\node (sa) at ($(llf.east) + (2.4,0)$) [anchor=west, draw=black, text width=0.3cm, align=center, inner sep=10pt] {\scriptsize SA};
\draw[-latex] (llf.east) ++(1.9,0) -- (sa);

\node (sum) at (upsamp |- sa) [circle, draw] {\Large $+$};
\draw[-latex] (sa) -- (sum);
\draw[-latex] (upsamp) -- (sum) ;

\draw[-latex] (upsamp) ++ (0,1) -| (sa);

\begin{scope}[shift={($(sum.east)+(0.8cm,0.4cm)$)}]
	\cube{0.5}{0.8}{0.8}{magentabox};
	\node at (0.7,0.5) [anchor=east] {\scriptsize $256\times 256 \times 128$};
\end{scope}

\begin{scope}[shift={($(sum.east)+(2.8cm,0.4cm)$)}]
	\cube{0.5}{0.8}{0.8}{magentabox};
	\node at (0.7,0.5) [anchor=east] {\scriptsize $256\times 256 \times 1$};
\end{scope}

\draw[-latex] (sum.east) -- ++(0.3,0);
\draw[-latex] (sum.east) ++(0.9,0) -- ++(1.4,0) node[midway, below, text width=1cm, align=center, labelnode] {\scriptsize $3\times 3\times 1$ conv.};
\node (loss) at  ($(sum.east)+(2.6cm,-2cm)$) [draw=black, thick, text width = 0.8cm, align=center, inner sep=10pt] {Loss};

\draw[latex-] (loss.north) -- ++(0, 1.1);

\node (gt) at (loss |- ca) [draw=black, align=center, text width = 0.8cm, inner sep=10pt, labelnode] {\scriptsize Ground\\truth};
\draw[-latex] (gt) -- (loss) ;

\end{tikzpicture}
        \caption{System architecture: light field FEE block receives the light fields and computes features. Spatial attention block (SA) and channel-wise attention block (CA) receives low level (Conv 1-2 and Conv 2-2) and high level (Conv3-3,  Conv 4-3 and Conv 5-3) features, respectively. VGG-16, or a similar block, produces these feature maps. Note that light field processing happens only in the light field FEE block. CA block gives attention to more informative kernel outputs. CPFE: context aware pyramid feature extraction.}\label{fi:architecture}
    \end{center}
    \vspace{-4ex}
\end{figure*}

Speeding-up light-field saliency detection requires avoiding computationally heavy one or more backbones and predominantly working in bulky light-field features maps. On the other hand, inability to freely augment light field images prevent training deep light field saliency detectors from scratch. These constraints demand using a pre-trained network (of course, followed by fine tuning). There are well-known pre-trained  networks that detect saliency in 2-D RGB images~\cite{zhao2019pyramid,SCRNet,GCPA}. In this paper we propose a FEE module that can be integrated into 2-D saliency detectors without any architectural changes to the \emph{base model}, to extract and encode the features in light fields. Figure~\ref{fi:architecture} shows an overview of the architecture of our system. The input to this neural network is a light field of size $S\times T\times U\times V$ in the form of a micro-lens image array of of size $W\times H$, where $W =S\times U$ and $H = T\times V$. Here, $(S,T)$ denotes the spatial resolution and $(U,V)$ denotes the angular resolution of a light field. Figure~\ref{fi:LF} shows a light field with sub-aperture images and micro lens array image. Then the extracted feature maps can be fed into the 2-D saliency detector to get the saliency maps. This whole network can be trained end-to-end manner after the integration.

\begin{figure*}[t!]
    \begin{center}
        \input{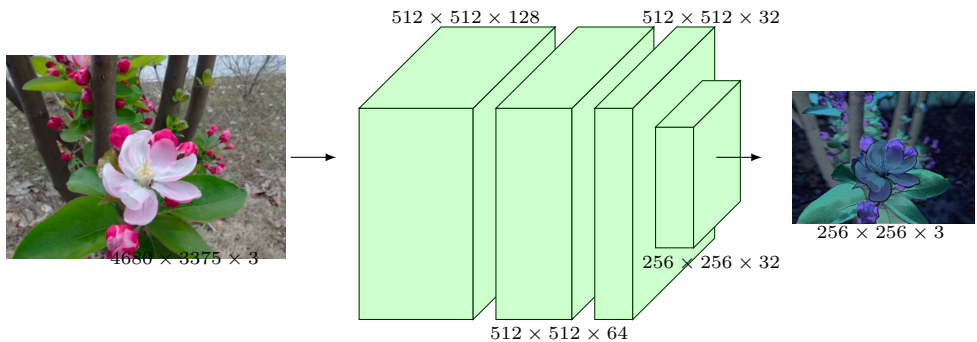}
        \caption{The light field FEE block receives micro lens array images of resolution $4608 \times 4608 \times 3$ and encodes the light field image into an RGB image of resolution $256\times 256\times 3$ through five convolutional layers.The FEE module consists of five convolutional layers. The first convolutional layer consist of $128$ filters of size $9\times 9$ and a $(9,9)$ stride. This layer precedes two convolutional layers having $64$ and $32$ filters of size $3\times 3$ and a stride $(1,1)$, respectively. A similar layer having $32$ filters of size $3\times 3$ and a stride $(2,2)$ is used to downsample the feature maps into $(256,256,32)$ and the last convolution layer has $3$ filters of size $1\times 1$ and a stride $(1,1)$.}
        \label{fi:FEE_block}
    \end{center}
    \vspace{-4ex}
\end{figure*}

\vspace{-1ex}
\subsection{2-D Saliency Detector}
\label{ssec:base_model}
Task of saliency detection in regular images is similar to binary semantic segmentation, and for this task requires both high level contextual information and low level spatial structural information. However, all of the high-level and low-level features are not suitable for saliency detection, and some features might even cause interference~\cite{zhao2019pyramid}. An attention mechanism can avoid such situations. The 2-D saliency detector proposed in~\cite{zhao2019pyramid} is such a system which we select as the saliency detector. This work especially uses channel wise attention module (CA) for high-level feature maps and spatial attention (SA) module for low-level feature maps with edge preserving loss function to preserve the edges of a saliency map. Along with the CA and SA modules, the pyramid feature network of the architecture leads to the state-of-the-art accuracy for RGB image saliency detection. However, using a single sub-aperture image or the all-focus image of a light field to feed the input of a 2-D saliency detector is ineffective as angular information of the light field gets lost. We solve this problem by using a carefully designed novel light field FEE module integrated in to the input of the network. We do not describe the architecture of the 2-D saliency detector in detail, and we refer the reader to~\cite{zhao2019pyramid} for more details.

\subsection{Novel Feature Extraction and Encoding Module}
\label{ssec:fe_block}
The 2-D saliency detector accepts inputs with resolution of $256\times256\times3$ and produces saliency maps with resolution of $256\times 256\times1$. Starting from this, our FEE module must extract and encode the pixel-wise angular information stored in a light field and produce an RGB image. In order to do that, by arranging a light field as a 2-D image of size $W\times H$, we run an $U\times V$ kernel with the stride of $(U,V)$ to exploit the angular information related to each pixel as mentioned in~\cite{zhang2020lightfield}. Here, we consider the modified light fields in the LYTRO ILLUM~\cite{zhang2020lightfield}, where $(U,V) = (9,9)$ and $(S,T) = (512,375)$ leading to $W = 4608$ and $H=3375$, and DUTLF-v2~\cite{DUTLF} ,where $(U,V) = (9,9)$ and $(S,T) = (512,400)$ leading to $W = 4608$ and $H=3600$. \emph{Because our light filed saliency detector shown in Figure~\ref{fi:architecture} processes light fields only in the FEE module and prevents subsequent processing in the 2-D saliency detector, we can achieve significant saving of computational time.}

The FEE module as depicted in Figure~\ref{fi:FEE_block} is the key component that leads to significant speed improvements. The FEE module aggressively downsamples a light field and encodes features suitable to be fed to a regular CNN. The FEE module consists of five convolutional layers. The first convolutional layer consist of $128$ filters of size $9\times 9$ and a $(9,9)$ stride. This layer precedes two convolutional layers having $64$ and $32$ filters of size $3\times 3$ and a stride $(1,1)$, respectively. A similar layer having $32$ filters of size $3\times 3$ and a stride $(2,2)$ is used to downsample the feature maps into $(256,256,32)$ and the last convolution layer has $3$ filters of size $1\times 1$ and a stride $(1,1)$ and compute the encoded output that is fed to the 2-D RGB saliency detector. 

As the input light field is a micro lens array image, adjacent pixels in the first $9\times9$ block comprises the first pixel of each of the $81$ sub-aperture images, see Figure~\ref{fi:LFc}. Therefore, by using a stride of $(9,9)$ in the first convolutional layer, we capture the same pixel for all the sub-aperture images at each convolution step. Following this, we select the layer-size parameters of hidden convolutional layers to be compatible with the architecture of VGG-16 network with decreasing number of filters at each layer to encode the light field in to a feature map of $256\times256\times3$ resolution. We note that VGG-16 is just one choice of the back bone, and other backbones, e.g., ResNets are also suitable.

\section{Experimental Results}
\label{sec:results}
We present experimental results in this section. We employ the LYTRO ILLUM~\cite{zhang2020lightfield} and DUTLF-v2~\cite{DUTLF} datasets in the experiments with a computing platform comprising of an Intel Core i9-9900K (3.60 GHz) CPU, $32$ GB RAM and Nvidia RTX-2080Ti GPU. Note that even though two other light field saliency datasets, HFUT-Lytro~\cite{zhang2017saliency} and LFSD~\cite{li2014saliency}, are publicly available, they are not suitable for evaluation of our light field saliency detector due to the low angular resolution and unavailability of sub-aperture images. There are $640$ light fields in the LYTRO ILLUM dataset, and we compare the performance the proposed light field saliency detector with the state-of-the-art light field saliency detectors LFNet~\cite{zhang2020lightfield} and MTCN~\cite{zhang2020multi} and state-of-the-art 2-D saliency detectors NLDF~\cite{NDLF}, PAGE-Net~\cite{PAGE}, GCPA Net~\cite{GCPA}, SCRNet~\cite{SCRNet} and SODGAN~\cite{SODGAN} in terms of the accuracy achieved with five-fold cross validation and computational time. The DUTLF-v2 \cite{DUTLF} dataset contains 4208 light fields divided into test and train sets having 2961 and 1247 light fields, respectively. We compare the performance of the proposed method, fine-tuned in the train set, with state-of-the-art 2-D saliency detectors, DLFS~\cite{piao2019deep} and student networks proposed in~\cite{DUTLF} and light field saliency detectors, LFNet~\cite{zhang2020lightfield}, DisenFusion \cite{disenfusion}, ATAFNet \cite{ATAFNet}, CPFP \cite{CFPP}, MOLF~\cite{MoLF} and teacher network proposed in~\cite{DUTLF}.
\vspace{-1ex}
\subsection{Implementation and Training of the Proposed Light Field Saliency Detector}
\label{Implementation and training of the model}

To facilitate the proposed FEE module to encode a light field into $256\times256\times3$ feature map, we crop the initial micro lens array images of the LYTRO-ILLUM dataset\cite{zhang2020lightfield} of size $4860\times 3375\times 3$ into four images of size $4608\times3375\times3$, removing pixels at the borders. This leads to a dataset of $2560$ light fields, and we incorporate data augmentation, such as random rotations of $90^\circ$ and $180^\circ$, random brightness, saturation and contrast changing, and random shuffling of the colour channels without affecting the angular information available with a light field. For the DUTLF-v2 \cite{DUTLF}, we use the same configuration except cropping micro lens array images into an one $4608\times 3600\times 3$ micro lens array image. We train our saliency detector in three steps. First, we train 2-D saliency detector~\cite{zhao2019pyramid} on a combined dataset of DUTS-TR \cite{wang2017DUTS} and ECSSD \cite{ECSSD} datasets with DUTS-TE \cite{wang2017DUTS} as the test set because the trained model of 2-D RGB saliency detector~\cite{zhao2019pyramid} is not available. We use the best performing model with a mean absolute error (MAE) of $0.0698$ as the base model even though this best model does not achieve the performance measures reported in \cite{zhao2019pyramid}, i.e., (MAE$=0.0405$). Then, we train the FEE module with the overall architecture shown in Figure~\ref{fi:architecture} using the light field dataset with the 2-D saliency detector frozen for 10 epochs. Finally, we train both FEE module and the 2-D saliency detector for another 40 epochs. For all the training, we employ the SGD optimizer~\cite{pmlr-v28-sutskever13} with a momentum of $0.9$, decay of $0$, and  initial learning rate of $10^{-2}$ with a batch size $8$. We use the loss function used in~\cite{zhao2019pyramid}, i.e.,

\vspace{-1ex}
\begin{equation}
\label{eq:loss}
    L = - \sum_{i=1}^{B} (\alpha_{s}Y_{i}\log(P_{i}))+(1-\alpha_{s})(1-Y_{i})\log(1-P_{i}),
\end{equation}

where $P_{i}$ is the predicted saliency map, $Y_{i}$ is the ground truth saliency map, $B$ is the batch size, and $\alpha_{s} = 0.528$~\cite{zhao2019pyramid}.

\subsection{Comparison with State-of-the-Art Light Field Saliency Detectors}
\label{lf_model_comparison}

\begin{figure*}
     \centering
     \begin{subfigure}[b]{0.14\linewidth}
         \centering
         \includegraphics[width=0.95\linewidth]{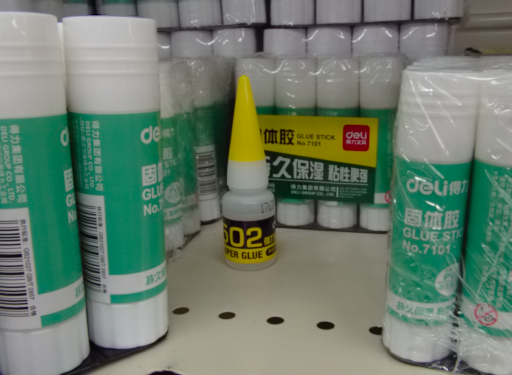}
     \end{subfigure}%
     \begin{subfigure}[b]{0.14\linewidth}
         \centering
         \includegraphics[width=0.95\linewidth]{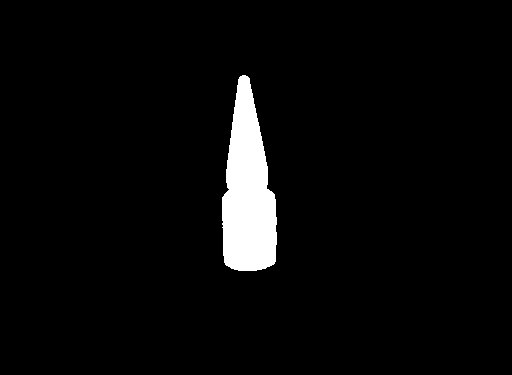}
     \end{subfigure}%
     \begin{subfigure}[b]{0.14\linewidth}
         \centering
         \includegraphics[width=0.95\linewidth]{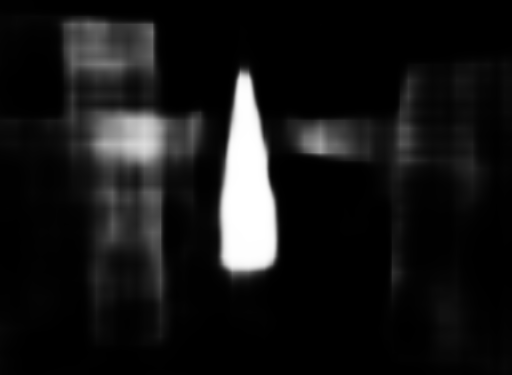}
     \end{subfigure}%
     \begin{subfigure}[b]{0.14\linewidth}
         \centering
         \includegraphics[width=0.95\linewidth]{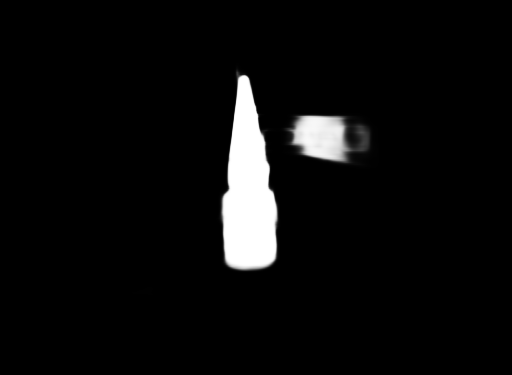}
     \end{subfigure}%
     \begin{subfigure}[b]{0.14\linewidth}
         \centering
         \includegraphics[width=0.95\linewidth]{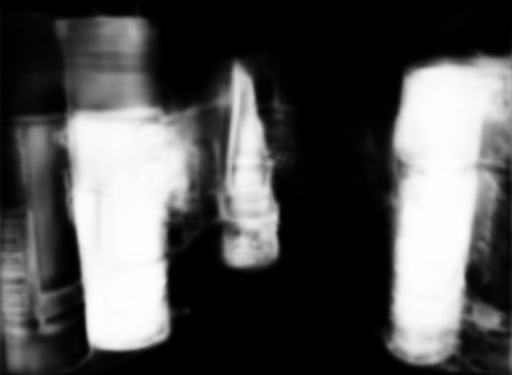}
     \end{subfigure}%
     \begin{subfigure}[b]{0.14\linewidth}
         \centering
         \includegraphics[width=0.95\linewidth]{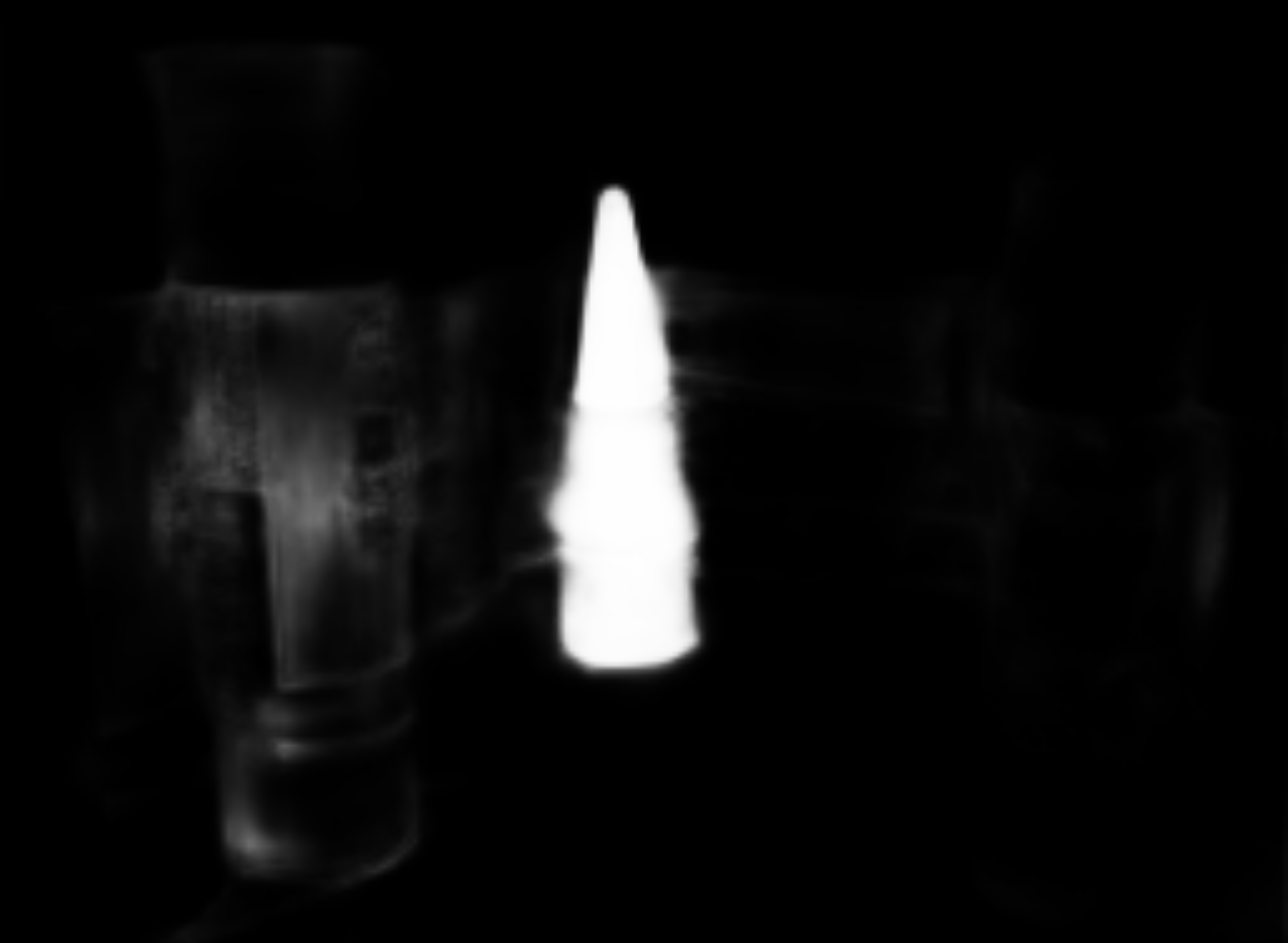}
     \end{subfigure}
     
     \begin{subfigure}[b]{0.14\linewidth}
         \centering
         \includegraphics[width=0.95\linewidth]{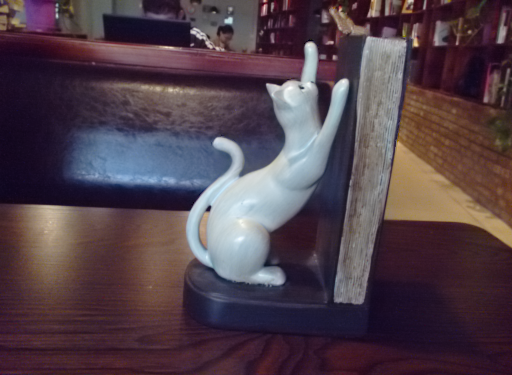}
     \end{subfigure}%
     \begin{subfigure}[b]{0.14\linewidth}
         \centering
         \includegraphics[width=0.95\linewidth]{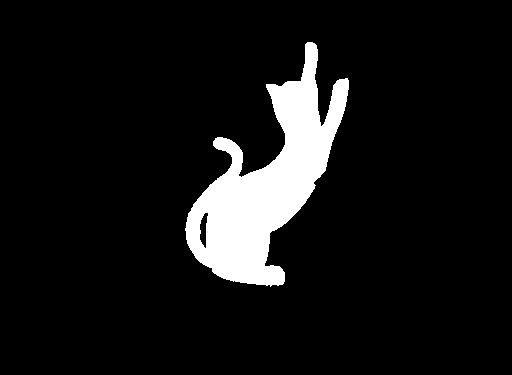}
     \end{subfigure}%
     \begin{subfigure}[b]{0.14\linewidth}
         \centering
         \includegraphics[width=0.95\linewidth]{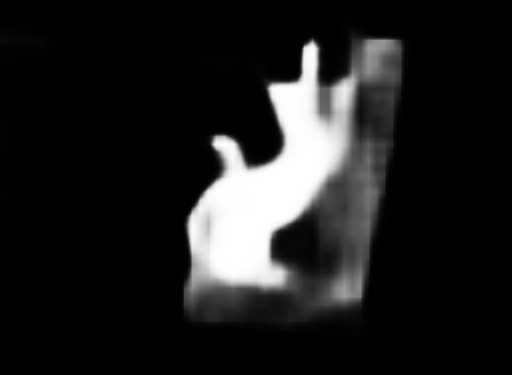}
     \end{subfigure}%
     \begin{subfigure}[b]{0.14\linewidth}
         \centering
         \includegraphics[width=0.95\linewidth]{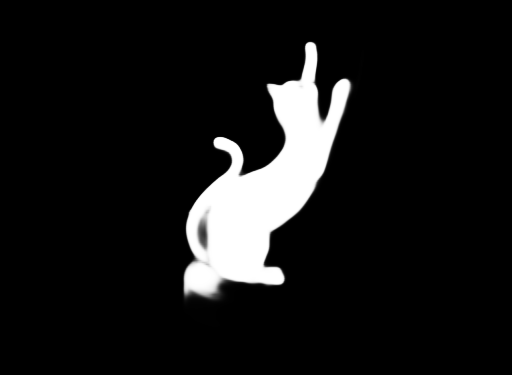}
     \end{subfigure}%
     \begin{subfigure}[b]{0.14\linewidth}
         \centering
         \includegraphics[width=0.95\linewidth]{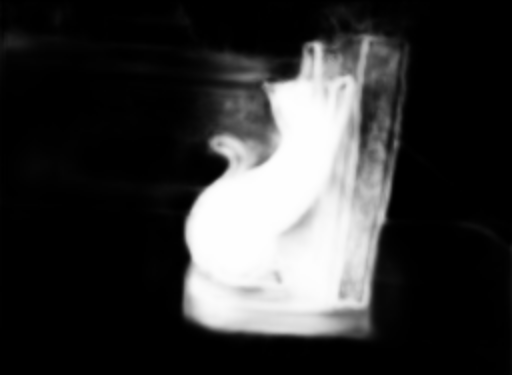}
     \end{subfigure}%
     \begin{subfigure}[b]{0.14\linewidth}
         \centering
         \includegraphics[width=0.95\linewidth]{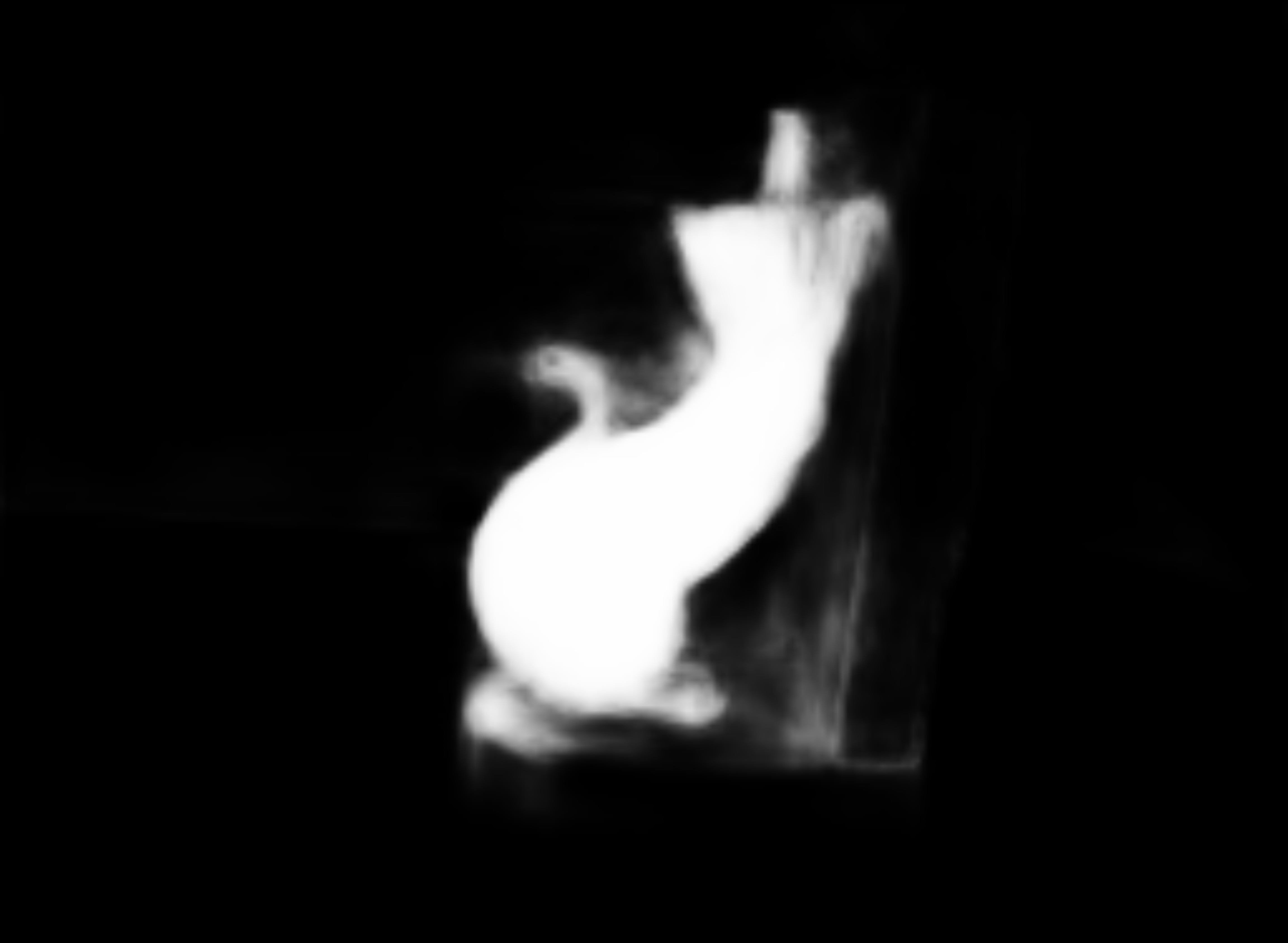}
     \end{subfigure}     

     \begin{subfigure}[b]{0.14\linewidth}
         \centering
         \includegraphics[width=0.95\linewidth]{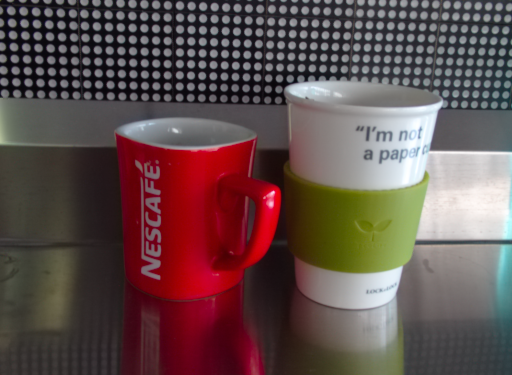}
     \end{subfigure}%
     \begin{subfigure}[b]{0.14\linewidth}
         \centering
         \includegraphics[width=0.95\linewidth]{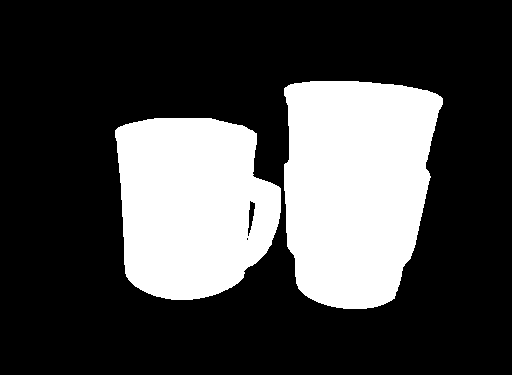}
     \end{subfigure}%
     \begin{subfigure}[b]{0.14\linewidth}
         \centering
         \includegraphics[width=0.95\linewidth]{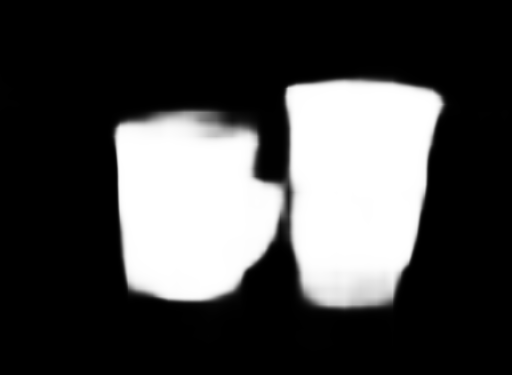}
     \end{subfigure}%
     \begin{subfigure}[b]{0.14\linewidth}
         \centering
         \includegraphics[width=0.95\linewidth]{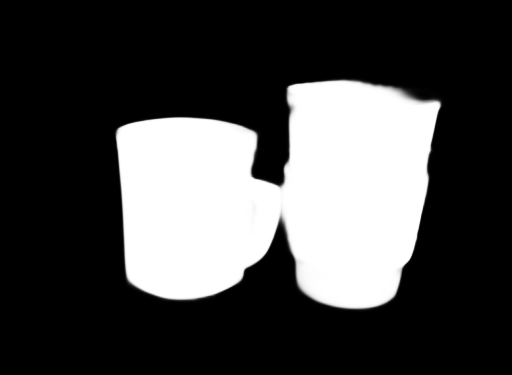}
     \end{subfigure}%
     \begin{subfigure}[b]{0.14\linewidth}
         \centering
         \includegraphics[width=0.95\linewidth]{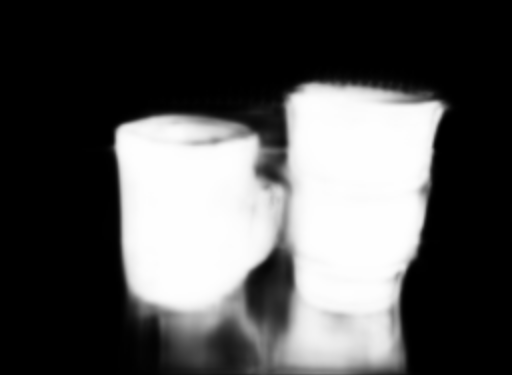}
     \end{subfigure}%
     \begin{subfigure}[b]{0.14\linewidth}
         \centering
         \includegraphics[width=0.95\linewidth]{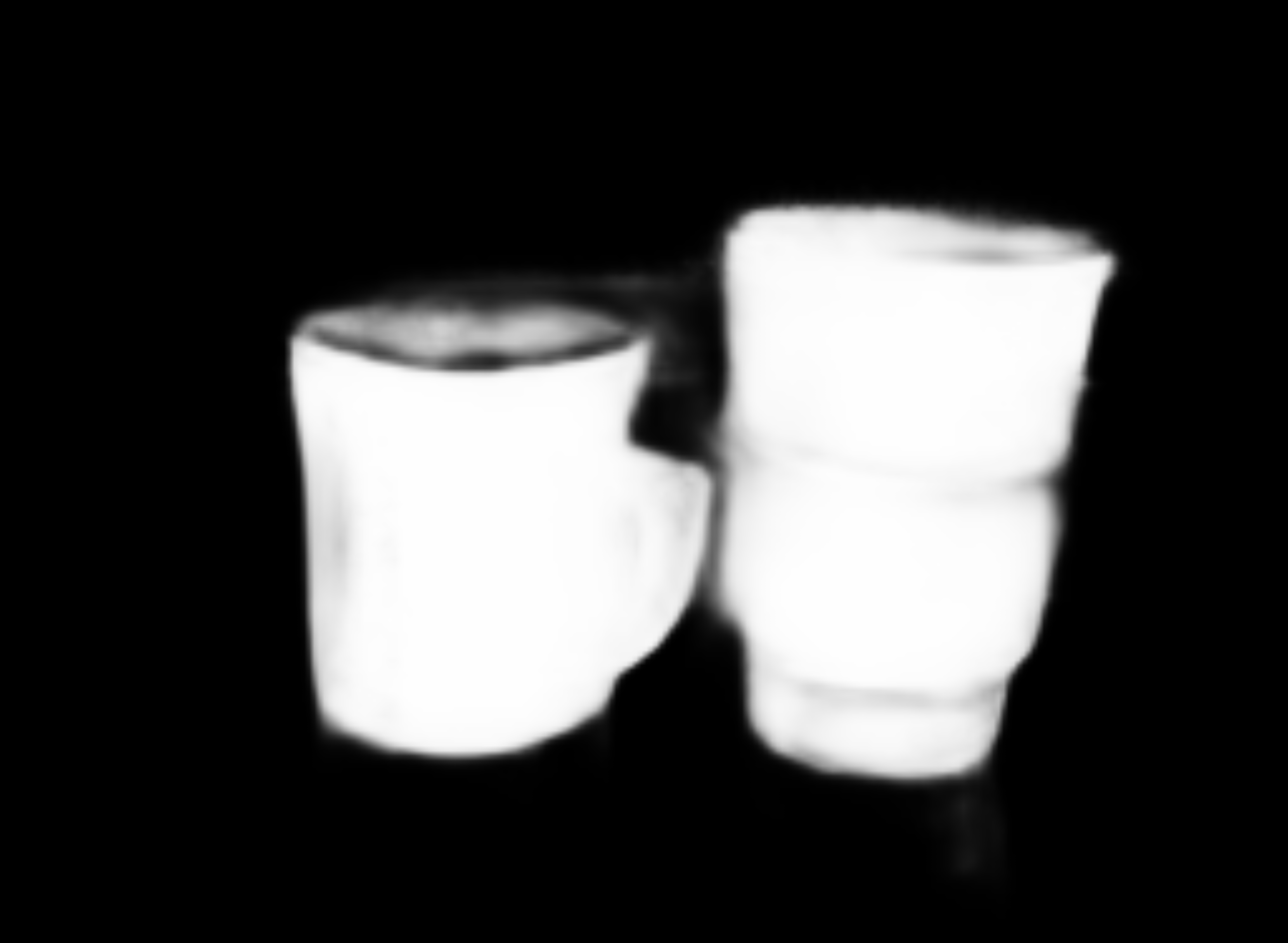}
     \end{subfigure}
     
     \begin{subfigure}[b]{0.14\linewidth}
         \centering
         \includegraphics[width=0.95\linewidth]{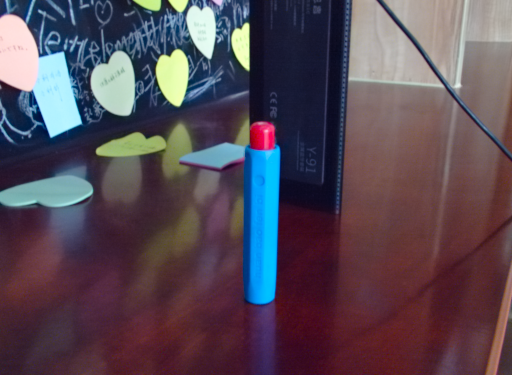}
     \end{subfigure}%
     \begin{subfigure}[b]{0.14\linewidth}
         \centering
         \includegraphics[width=0.95\linewidth]{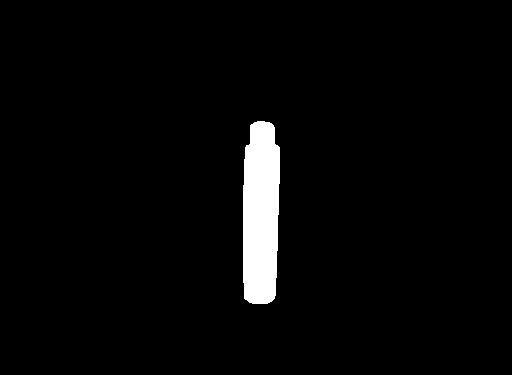}
     \end{subfigure}%
     \begin{subfigure}[b]{0.14\linewidth}
         \centering
         \includegraphics[width=0.95\linewidth]{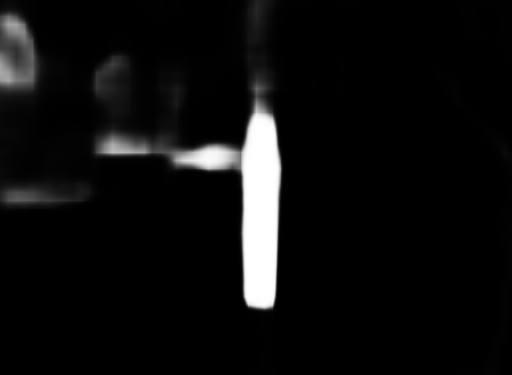}
     \end{subfigure}%
     \begin{subfigure}[b]{0.14\linewidth}
         \centering
         \includegraphics[width=0.95\linewidth]{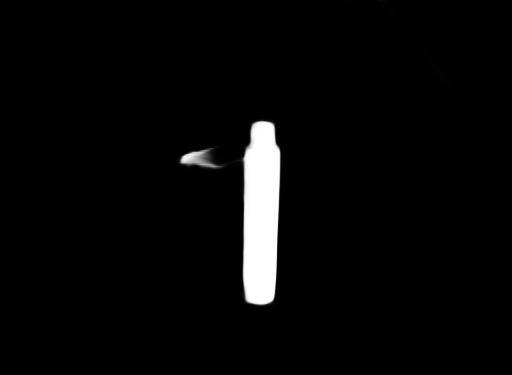}
     \end{subfigure}%
     \begin{subfigure}[b]{0.14\linewidth}
         \centering
         \includegraphics[width=0.95\linewidth]{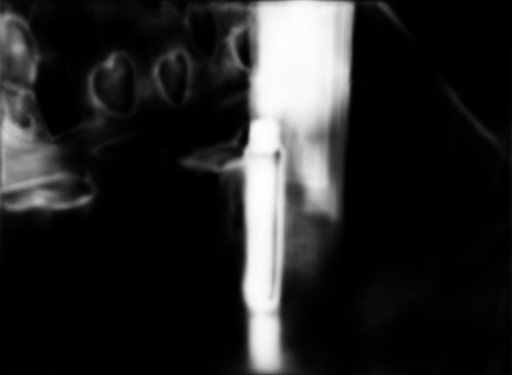}
     \end{subfigure}%
     \begin{subfigure}[b]{0.14\linewidth}
         \centering
         \includegraphics[width=0.95\linewidth]{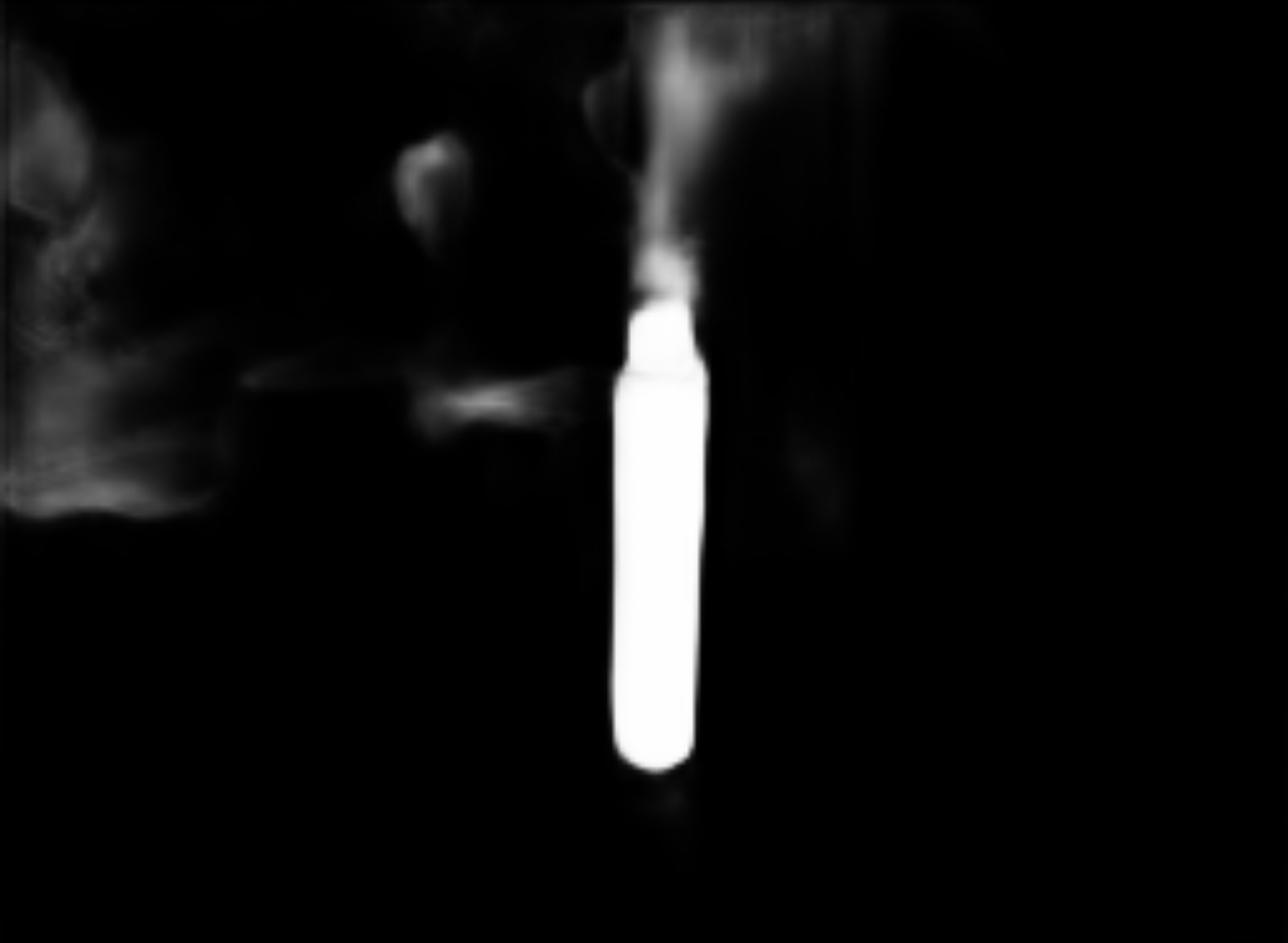}
     \end{subfigure}
     
    \begin{subfigure}[b]{0.14\linewidth}
         \centering
         \includegraphics[width=0.95\linewidth]{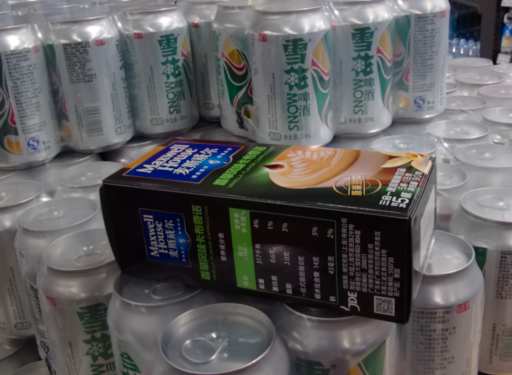}
     \end{subfigure}%
     \begin{subfigure}[b]{0.14\linewidth}
         \centering
         \includegraphics[width=0.95\linewidth]{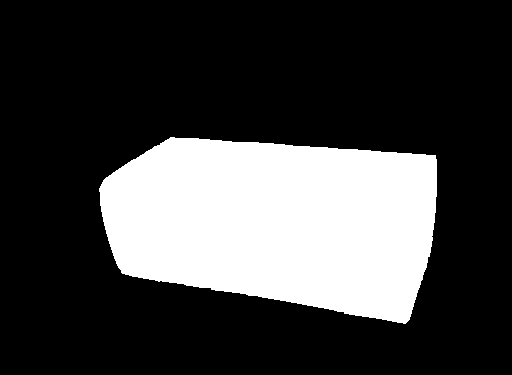}
     \end{subfigure}%
     \begin{subfigure}[b]{0.14\linewidth}
         \centering
         \includegraphics[width=0.95\linewidth]{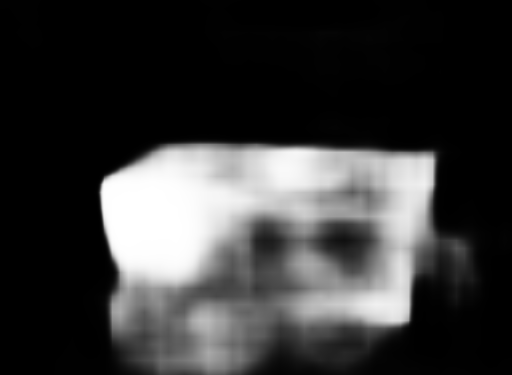}
     \end{subfigure}%
     \begin{subfigure}[b]{0.14\linewidth}
         \centering
         \includegraphics[width=0.95\linewidth]{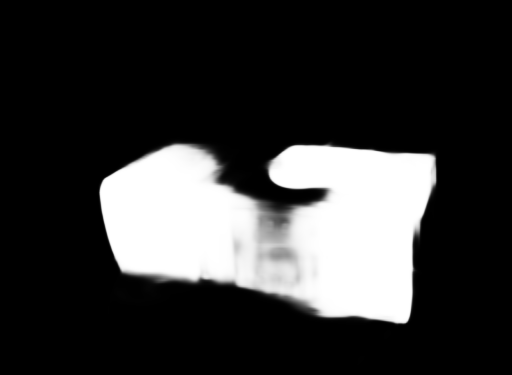}
     \end{subfigure}%
     \begin{subfigure}[b]{0.14\linewidth}
         \centering
         \includegraphics[width=0.95\linewidth]{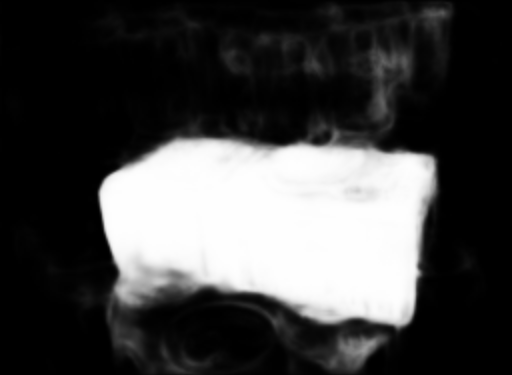}
     \end{subfigure}%
     \begin{subfigure}[b]{0.14\linewidth}
         \centering
         \includegraphics[width=0.95\linewidth]{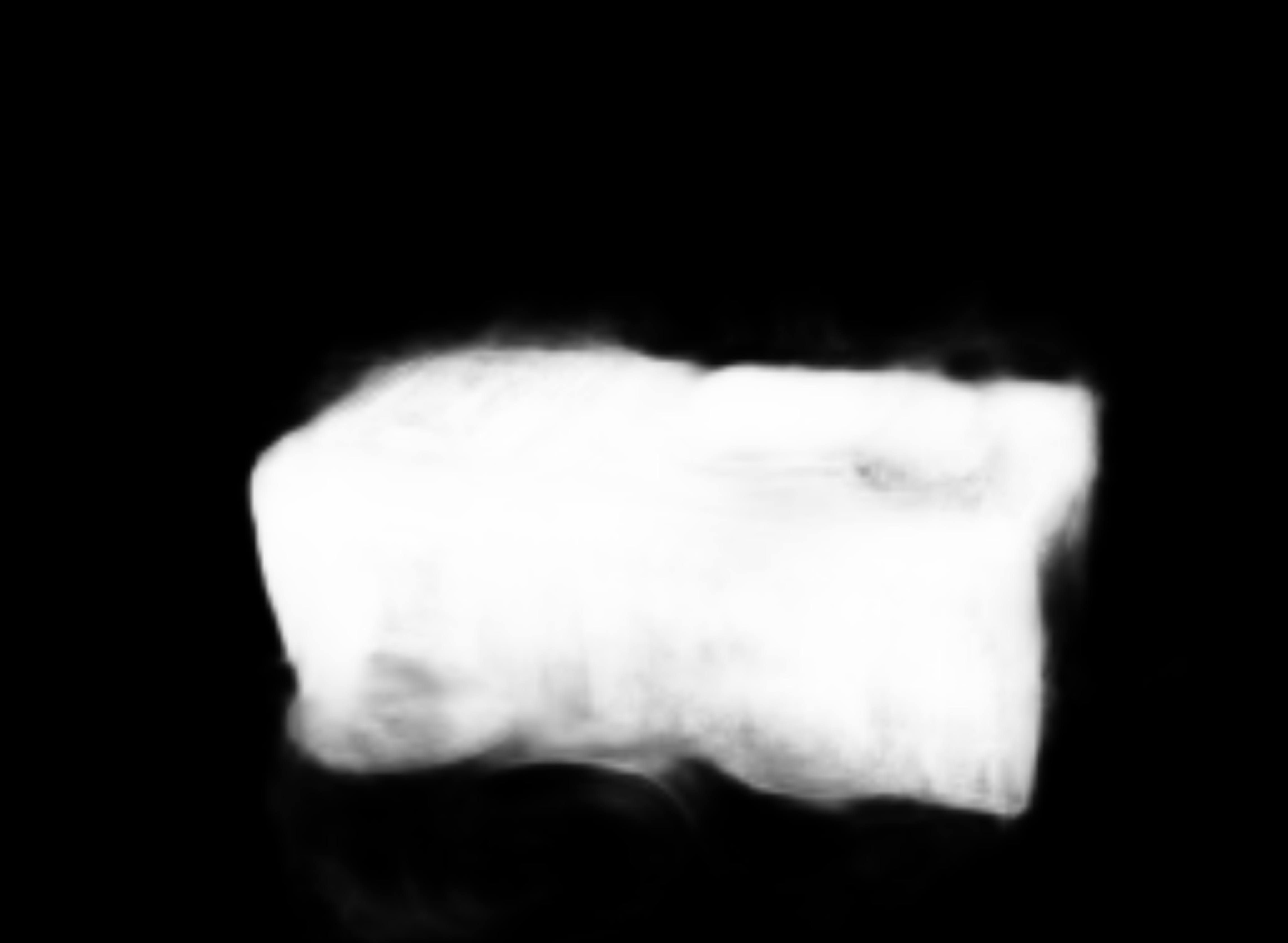}
     \end{subfigure}
     
     \begin{subfigure}[b]{0.14\linewidth}
         \centering
         \includegraphics[width=0.95\linewidth]{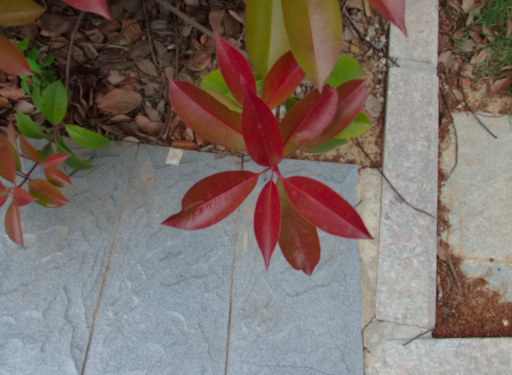}
     \end{subfigure}%
     \begin{subfigure}[b]{0.14\linewidth}
         \centering
         \includegraphics[width=0.95\linewidth]{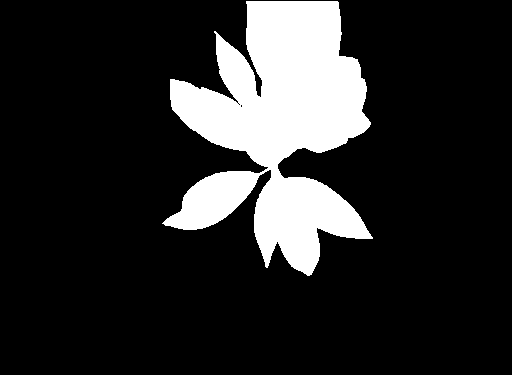}
     \end{subfigure}%
     \begin{subfigure}[b]{0.14\linewidth}
         \centering
         \includegraphics[width=0.95\linewidth]{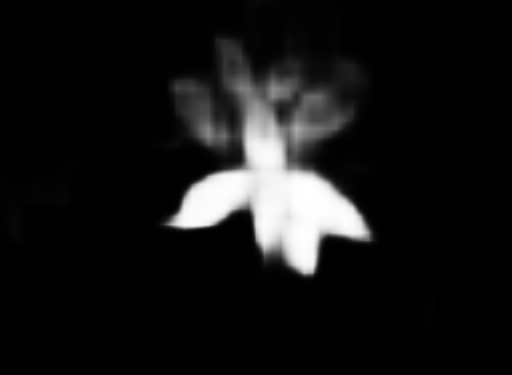}
     \end{subfigure}%
     \begin{subfigure}[b]{0.14\linewidth}
         \centering
         \includegraphics[width=0.95\linewidth]{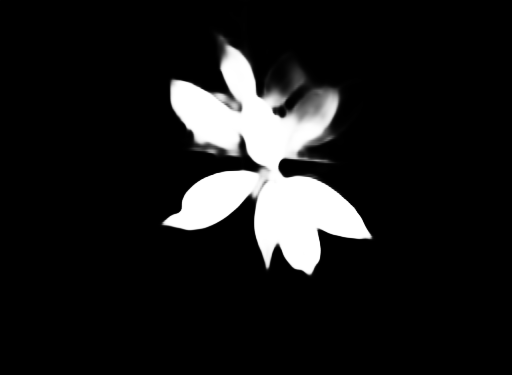}
     \end{subfigure}%
     \begin{subfigure}[b]{0.14\linewidth}
         \centering
         \includegraphics[width=0.95\linewidth]{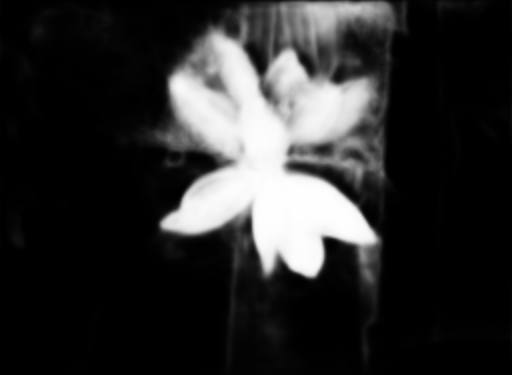}
     \end{subfigure}%
     \begin{subfigure}[b]{0.14\linewidth}
         \centering
         \includegraphics[width=0.95\linewidth]{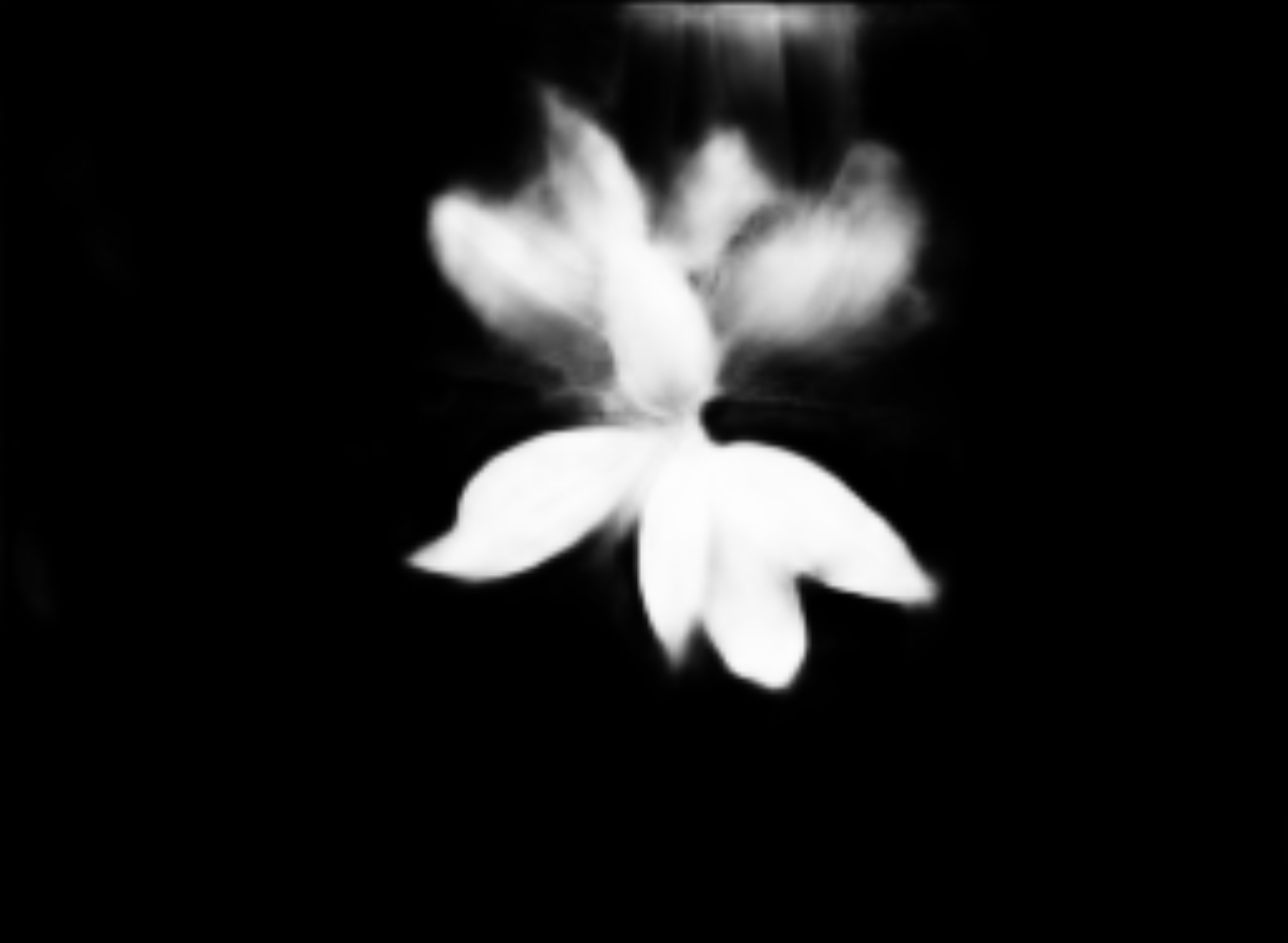}
     \end{subfigure}
     
     \begin{subfigure}[b]{0.14\linewidth}
         \centering
         \includegraphics[width=0.95\linewidth]{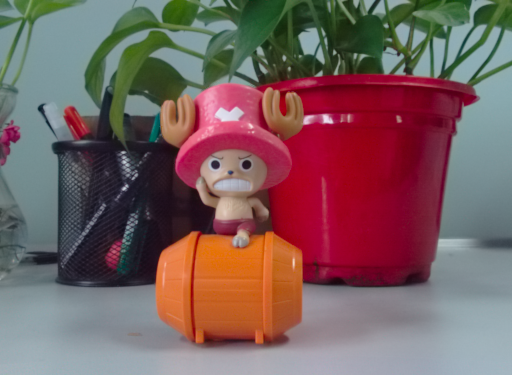}
     \end{subfigure}%
     \begin{subfigure}[b]{0.14\linewidth}
         \centering
         \includegraphics[width=0.95\linewidth]{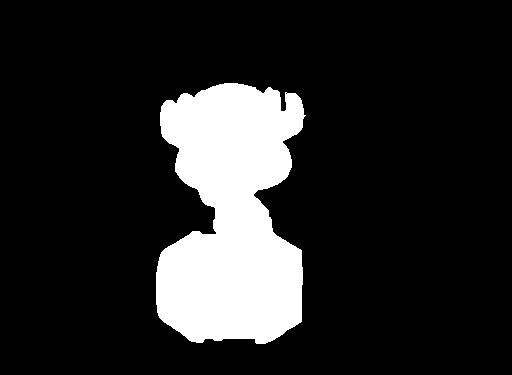}
     \end{subfigure}%
     \begin{subfigure}[b]{0.14\linewidth}
         \centering
         \includegraphics[width=0.95\linewidth]{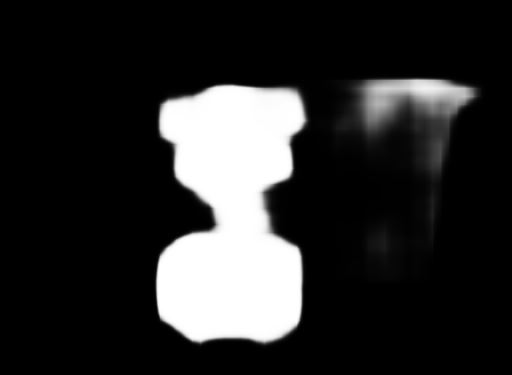}
     \end{subfigure}%
     \begin{subfigure}[b]{0.14\linewidth}
         \centering
         \includegraphics[width=0.95\linewidth]{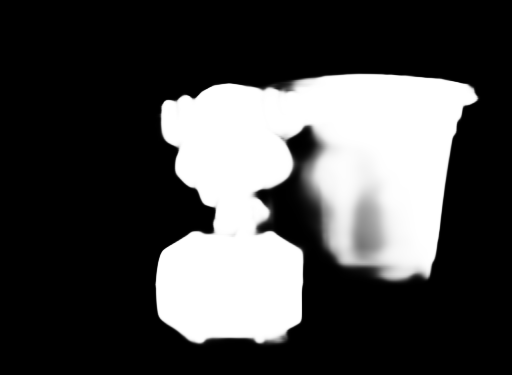}
     \end{subfigure}%
     \begin{subfigure}[b]{0.14\linewidth}
         \centering
         \includegraphics[width=0.95\linewidth]{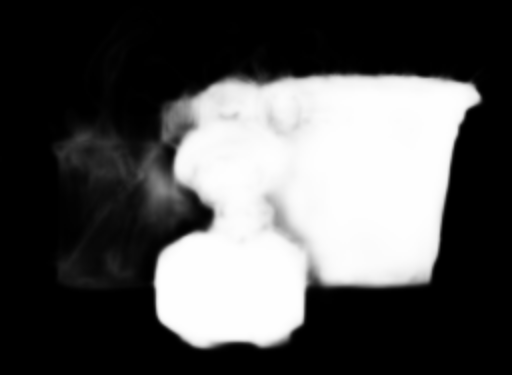}
     \end{subfigure}%
     \begin{subfigure}[b]{0.14\linewidth}
         \centering
         \includegraphics[width=0.95\linewidth]{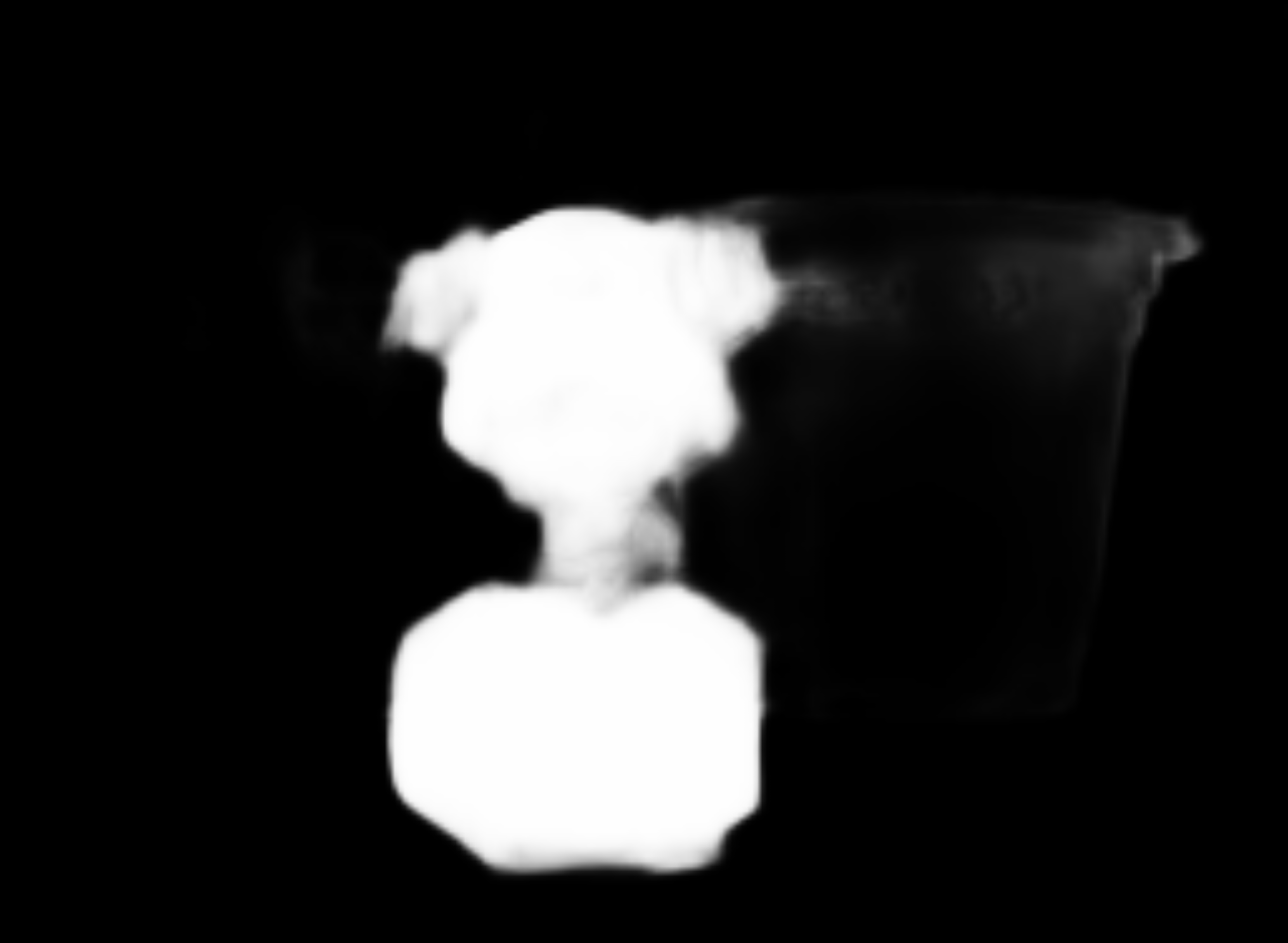}
     \end{subfigure}
     
     \begin{subfigure}[b]{0.14\linewidth}
         \centering
         \includegraphics[width=0.95\linewidth]{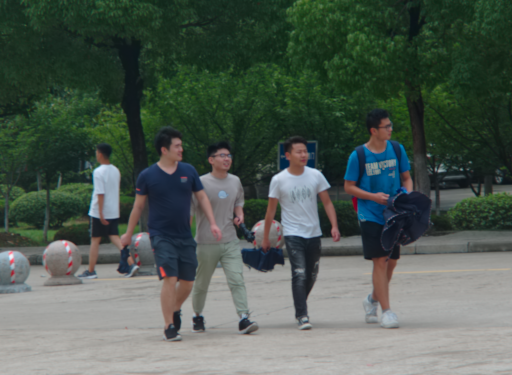}
     \end{subfigure}%
     \begin{subfigure}[b]{0.14\linewidth}
         \centering
         \includegraphics[width=0.95\linewidth]{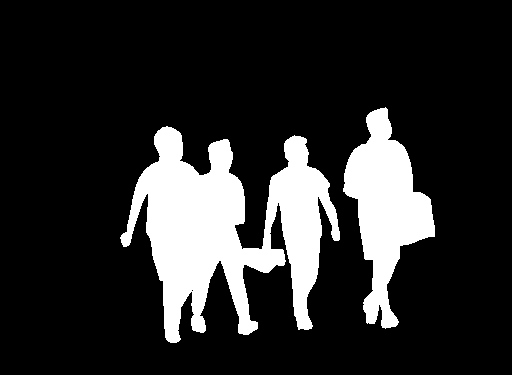}
     \end{subfigure}%
     \begin{subfigure}[b]{0.14\linewidth}
         \centering
         \includegraphics[width=0.95\linewidth]{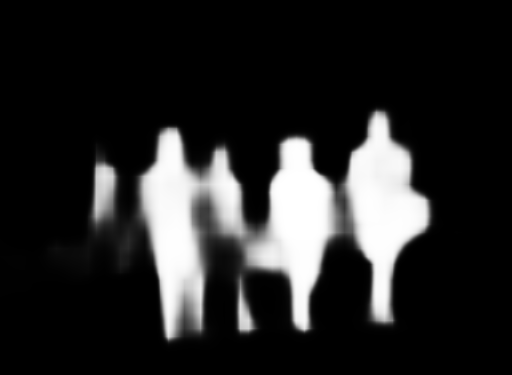}
     \end{subfigure}%
     \begin{subfigure}[b]{0.14\linewidth}
         \centering
         \includegraphics[width=0.95\linewidth]{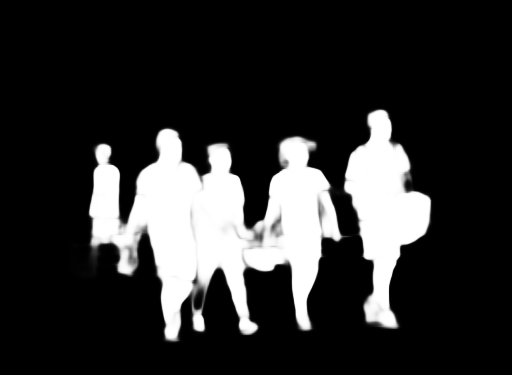}
     \end{subfigure}%
     \begin{subfigure}[b]{0.14\linewidth}
         \centering
         \includegraphics[width=0.95\linewidth]{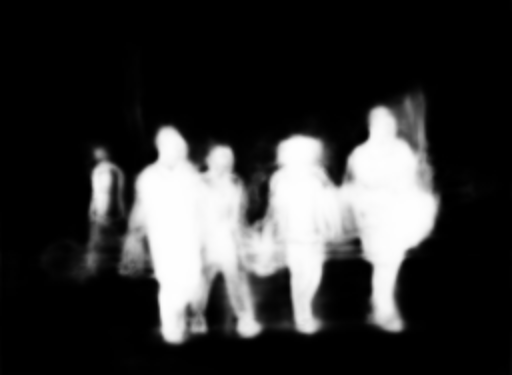}
     \end{subfigure}%
     \begin{subfigure}[b]{0.14\linewidth}
         \centering
         \includegraphics[width=0.95\linewidth]{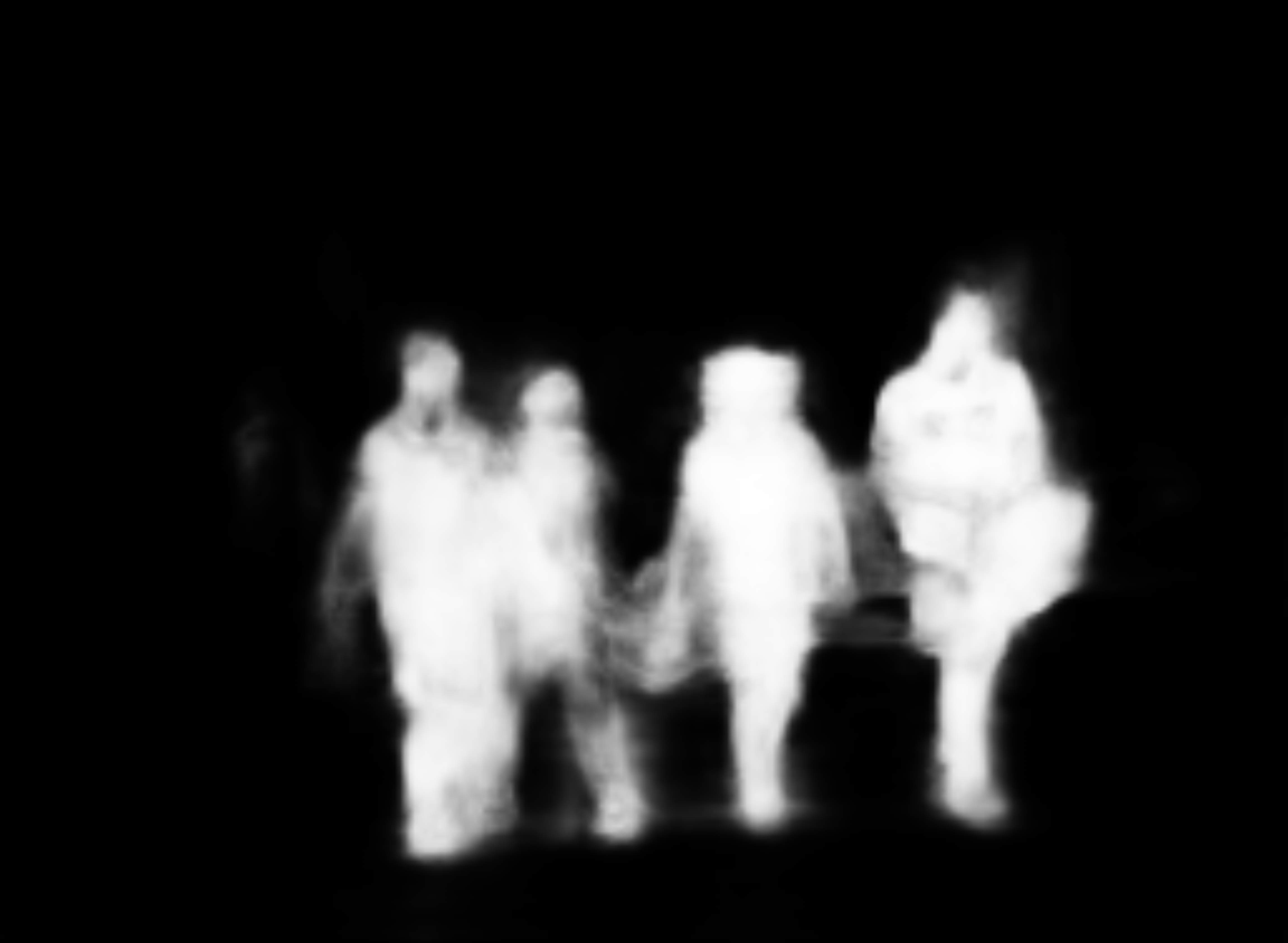}
     \end{subfigure}
     
     \begin{subfigure}[b]{0.14\linewidth}
         \centering
         \includegraphics[width=0.95\linewidth]{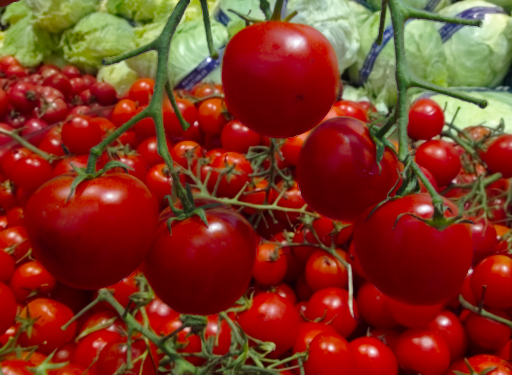}
     \end{subfigure}%
     \begin{subfigure}[b]{0.14\linewidth}
         \centering
         \includegraphics[width=0.95\linewidth]{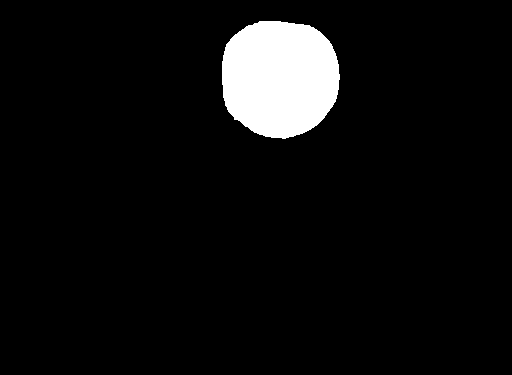}
     \end{subfigure}%
     \begin{subfigure}[b]{0.14\linewidth}
         \centering
         \includegraphics[width=0.95\linewidth]{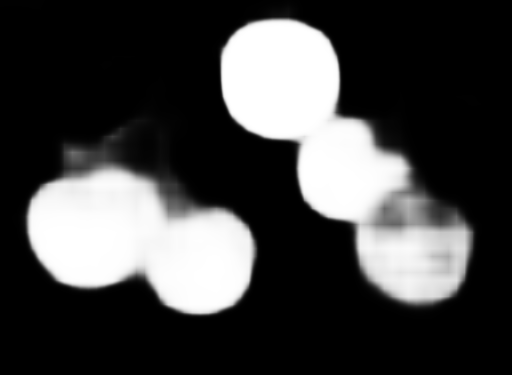}
     \end{subfigure}%
     \begin{subfigure}[b]{0.14\linewidth}
         \centering
         \includegraphics[width=0.95\linewidth]{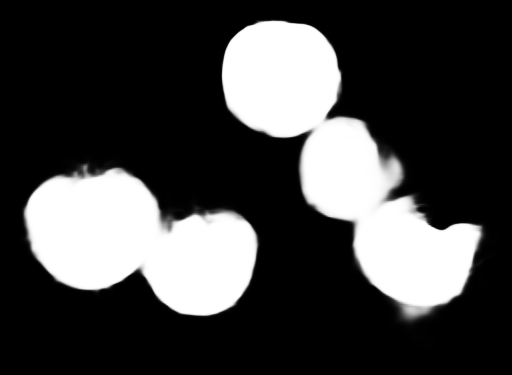}
     \end{subfigure}%
     \begin{subfigure}[b]{0.14\linewidth}
         \centering
         \includegraphics[width=0.95\linewidth]{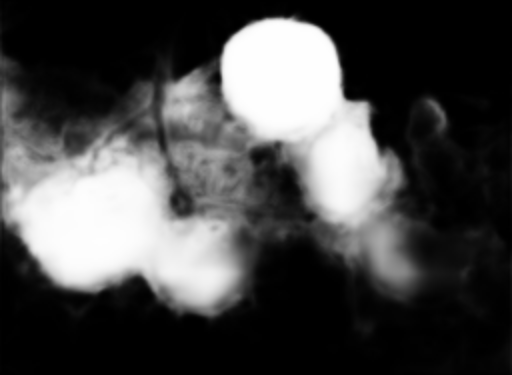}
     \end{subfigure}%
     \begin{subfigure}[b]{0.14\linewidth}
         \centering
         \includegraphics[width=0.95\linewidth]{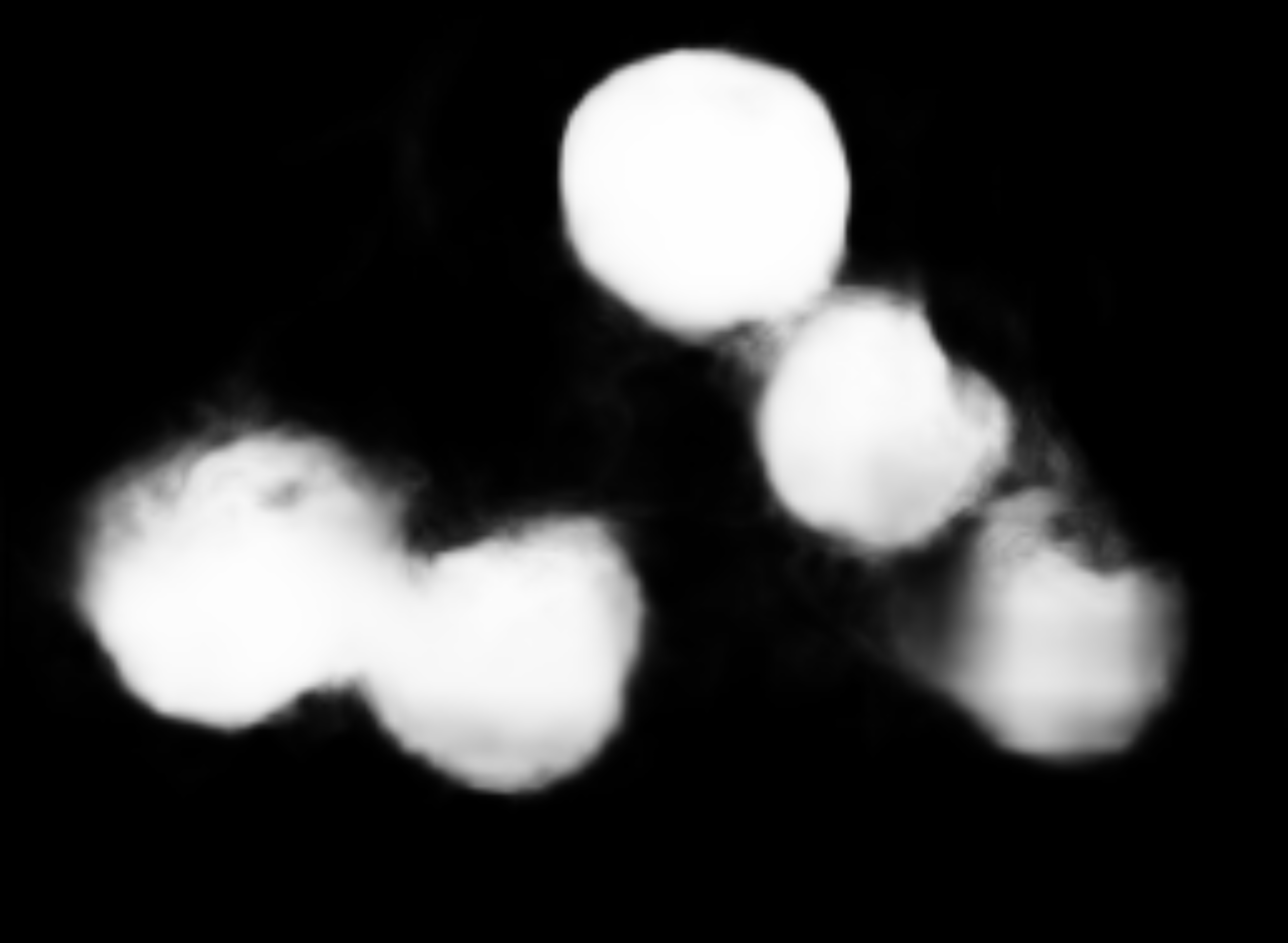}
     \end{subfigure}
     
     \begin{subfigure}[b]{0.14\linewidth}
         \centering
         \includegraphics[width=0.95\linewidth]{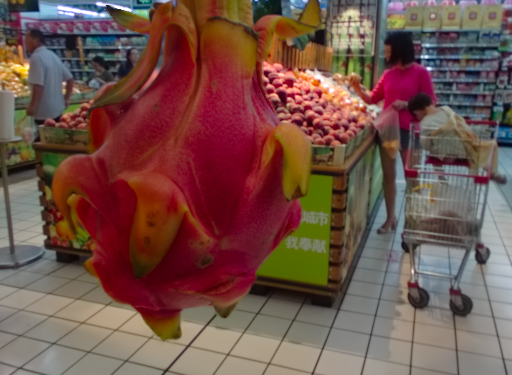}
     \end{subfigure}%
     \begin{subfigure}[b]{0.14\linewidth}
         \centering
         \includegraphics[width=0.95\linewidth]{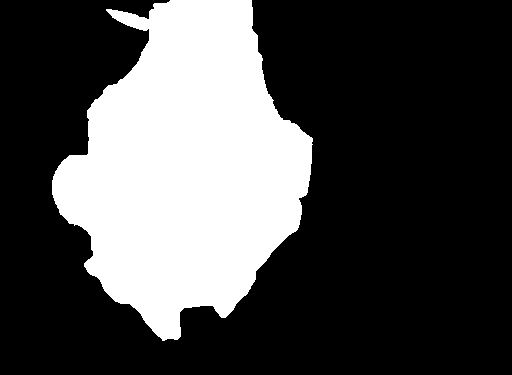}
     \end{subfigure}%
     \begin{subfigure}[b]{0.14\linewidth}
         \centering
         \includegraphics[width=0.95\linewidth]{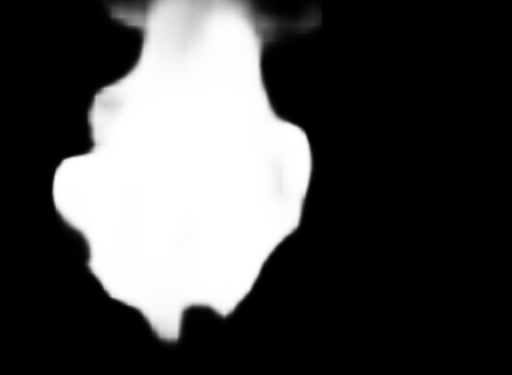}
     \end{subfigure}%
     \begin{subfigure}[b]{0.14\linewidth}
         \centering
         \includegraphics[width=0.95\linewidth]{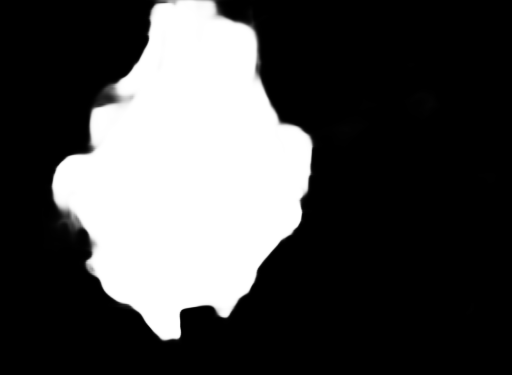}
     \end{subfigure}%
     \begin{subfigure}[b]{0.14\linewidth}
         \centering
         \includegraphics[width=0.95\linewidth]{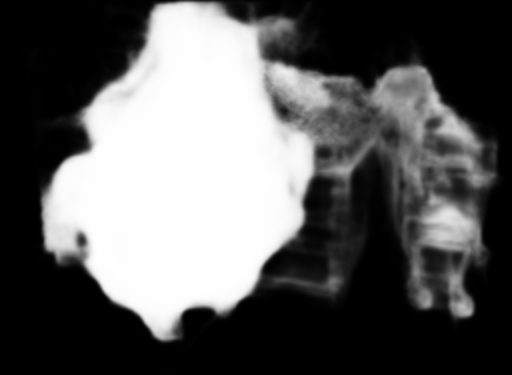}
     \end{subfigure}%
     \begin{subfigure}[b]{0.14\linewidth}
         \centering
         \includegraphics[width=0.95\linewidth]{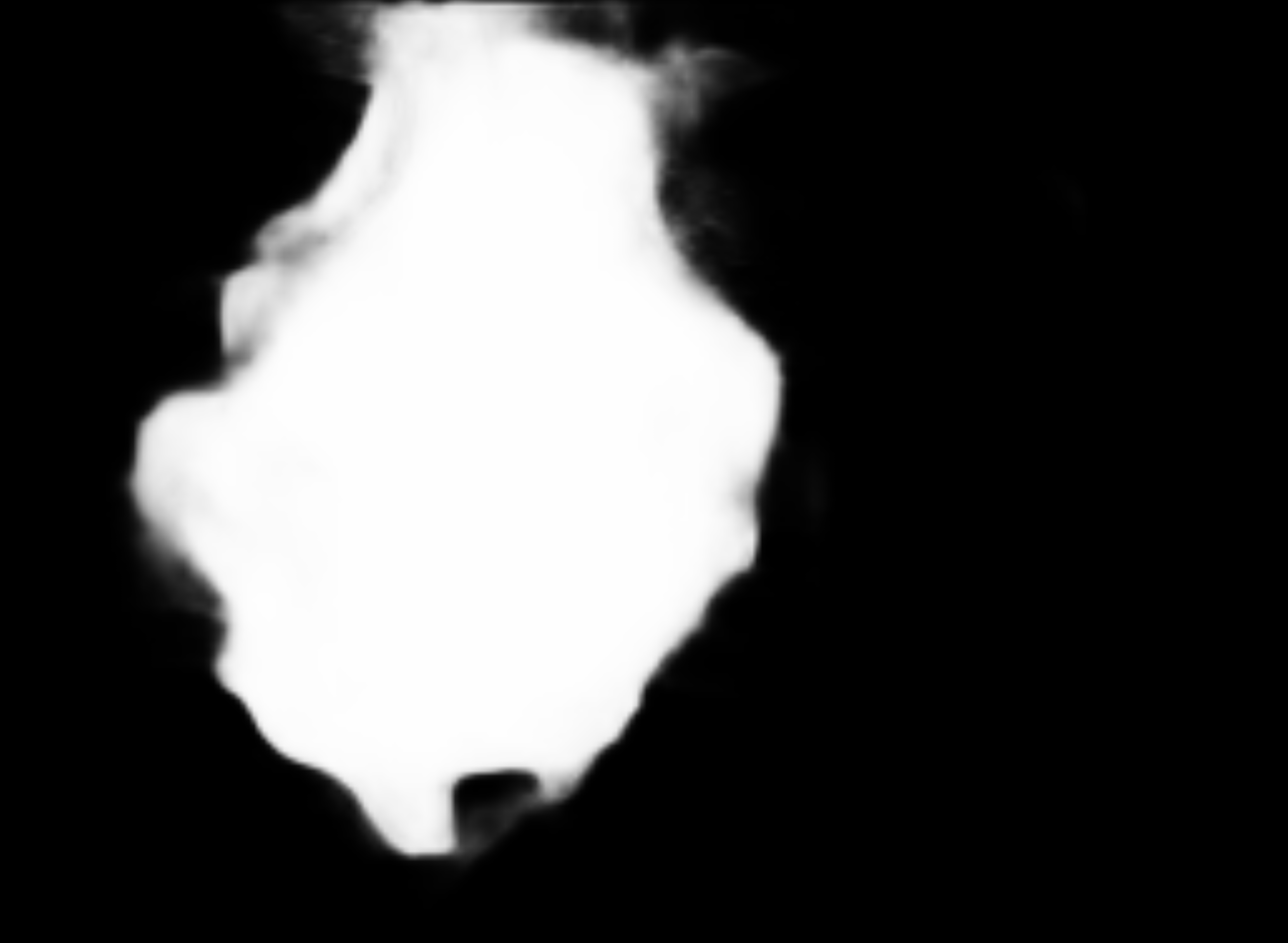}
     \end{subfigure}
     
     \begin{subfigure}[b]{0.14\linewidth}
         \centering
         \includegraphics[width=0.95\linewidth]{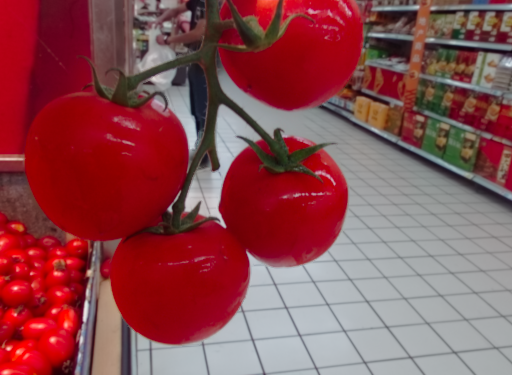}
     \end{subfigure}%
     \begin{subfigure}[b]{0.14\linewidth}
         \centering
         \includegraphics[width=0.95\linewidth]{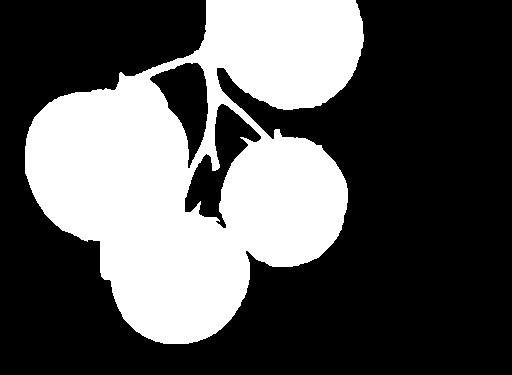}
     \end{subfigure}%
     \begin{subfigure}[b]{0.14\linewidth}
         \centering
         \includegraphics[width=0.95\linewidth]{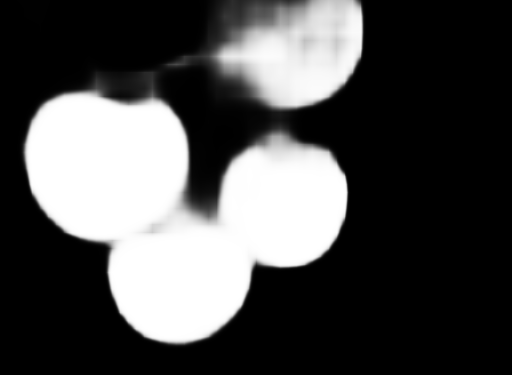}
     \end{subfigure}%
     \begin{subfigure}[b]{0.14\linewidth}
         \centering
         \includegraphics[width=0.95\linewidth]{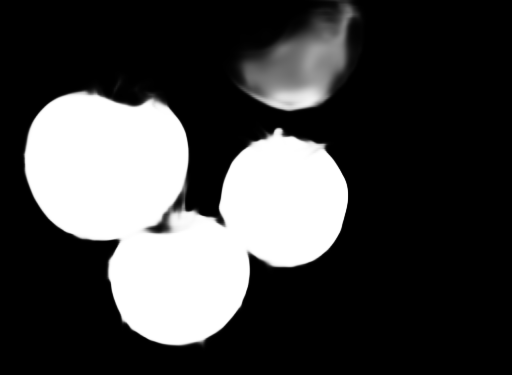}
     \end{subfigure}%
     \begin{subfigure}[b]{0.14\linewidth}
         \centering
         \includegraphics[width=0.95\linewidth]{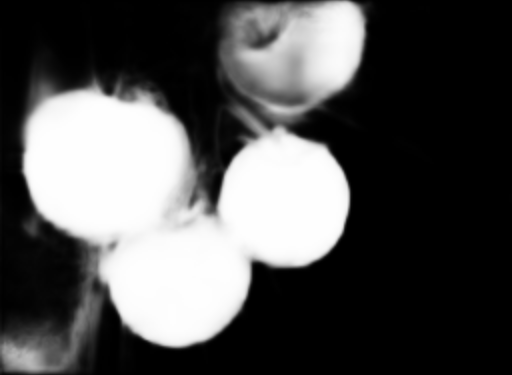}
     \end{subfigure}%
     \begin{subfigure}[b]{0.14\linewidth}
         \centering
         \includegraphics[width=0.95\linewidth]{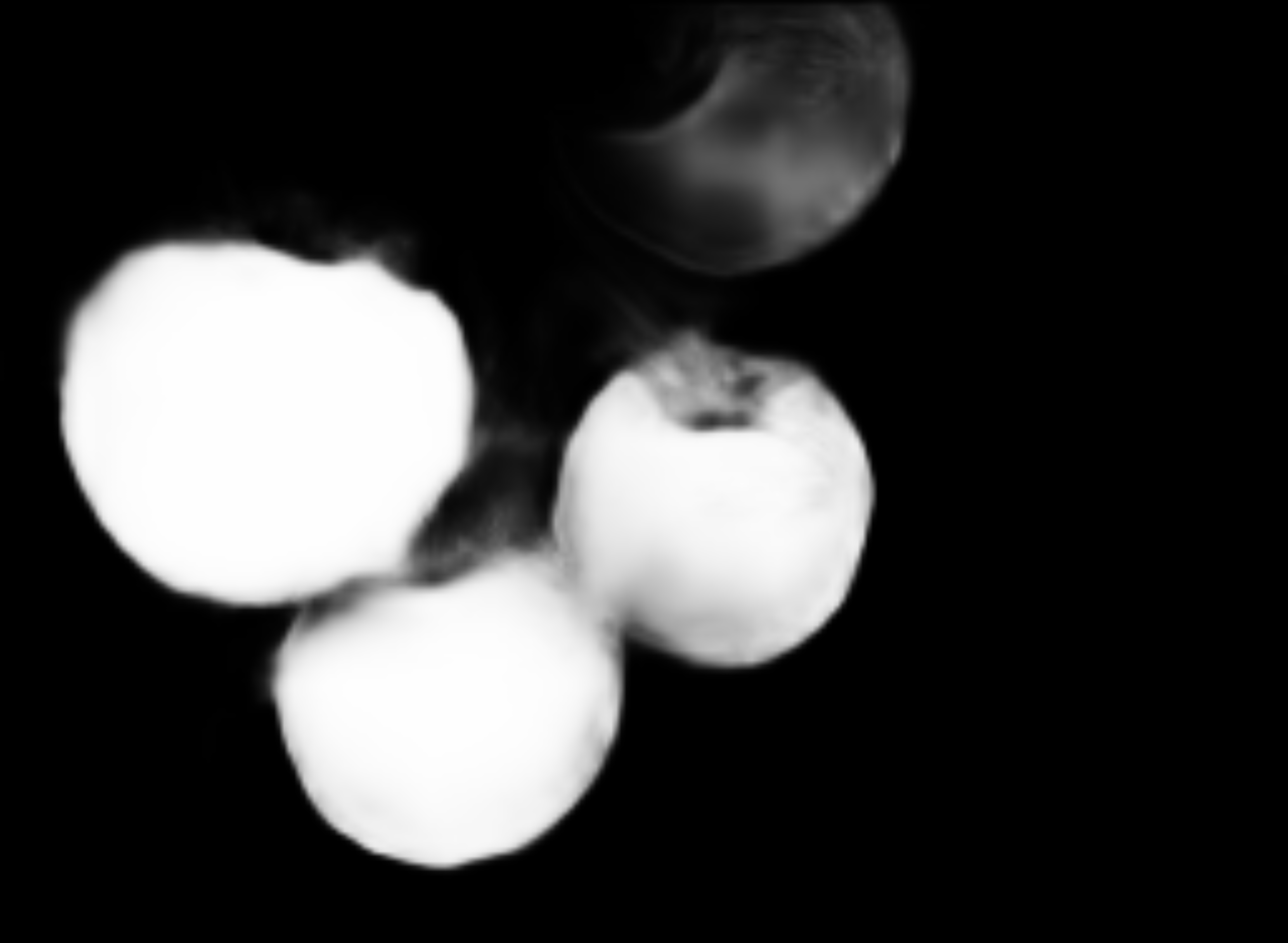}
     \end{subfigure}
     
     \begin{subfigure}[b]{0.14\linewidth}
         \centering
         \includegraphics[width=0.95\linewidth]{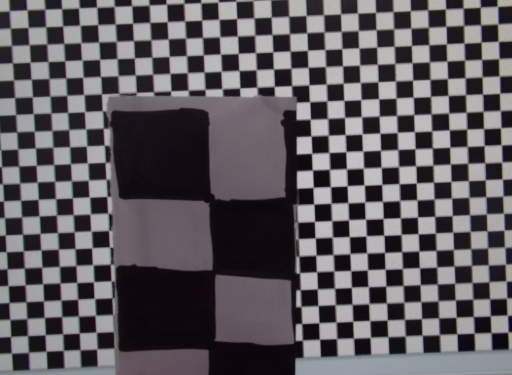}
         \caption{Center SAI}
         \label{sfi:sf1}
     \end{subfigure}%
     \begin{subfigure}[b]{0.14\linewidth}
         \centering
         \includegraphics[width=0.95\linewidth]{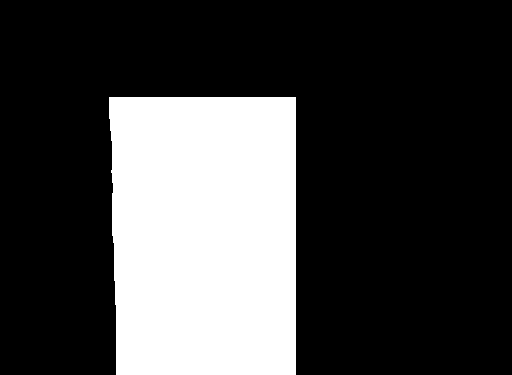}
         \caption{GT}
         \label{sfi:sf2}
     \end{subfigure}%
     \begin{subfigure}[b]{0.14\linewidth}
         \centering
         \includegraphics[width=0.95\linewidth]{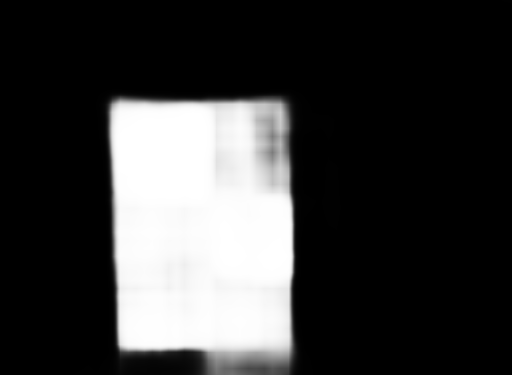}
         \caption{LFNet~\cite{zhang2020lightfield}}
         \label{sfi:sf3}
     \end{subfigure}%
     \begin{subfigure}[b]{0.14\linewidth}
         \centering
         \includegraphics[width=0.95\linewidth]{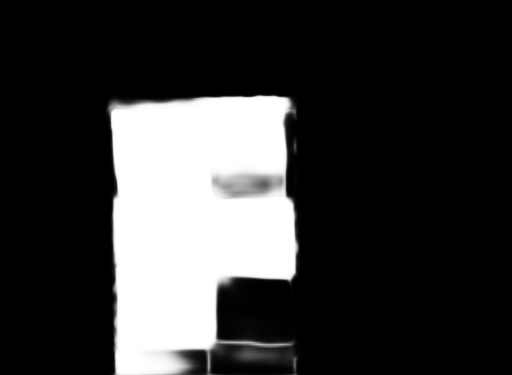}
         \caption{MTCN~\cite{zhang2020multi}}
         \label{sfi:sf3}
     \end{subfigure}%
     \begin{subfigure}[b]{0.14\linewidth}
         \centering
         \includegraphics[width=0.95\linewidth]{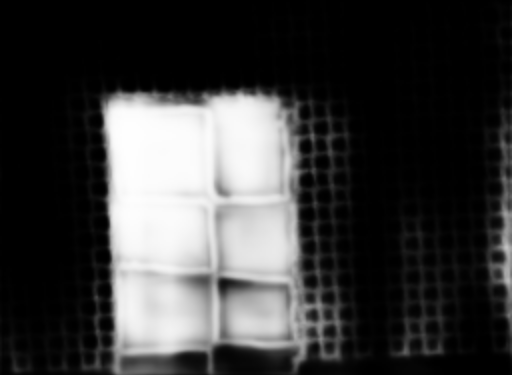}
         \caption{Base~\cite{zhao2019pyramid}}
         \label{sfi:sf3}
     \end{subfigure}%
     \begin{subfigure}[b]{0.14\linewidth}
         \centering
         \includegraphics[width=0.95\linewidth]{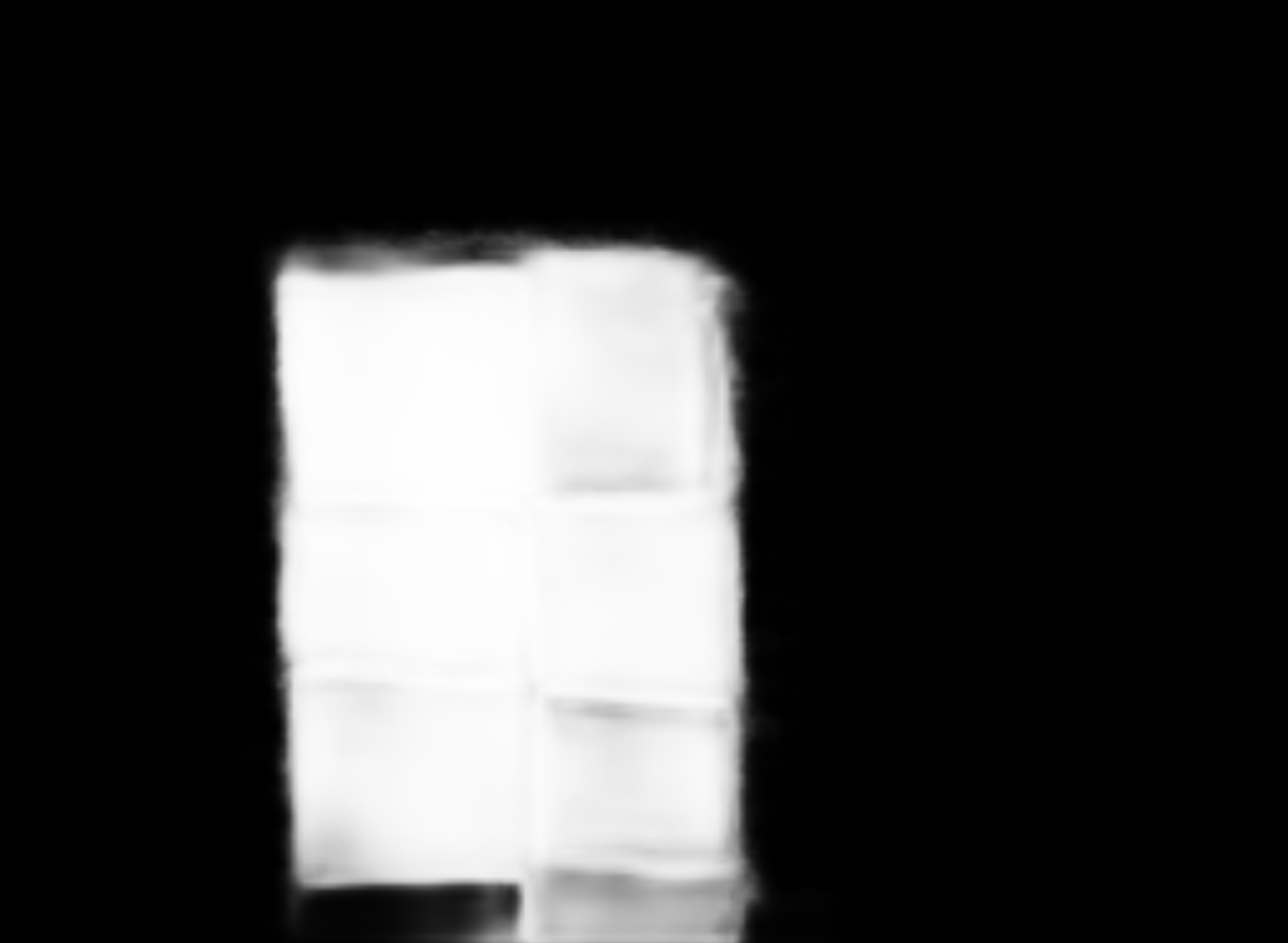}
         \caption{Ours}
         \label{sfi:sf4}
     \end{subfigure}     
    \caption{Comparison of saliency maps: (a) centre sub-aperture image (SAI) of the light field, (b) ground truth (GT), (c) LFNet results~\cite{zhang2020lightfield}, (d) our results. Our saliency maps are closer to the ground truth compared to those of LFNet\cite{zhang2020lightfield} and base model\cite{zhao2019pyramid}}
    \label{fi:pictorial_results}
  
\end{figure*}

\begin{figure}[H]
     \centering
     \begin{subfigure}[b]{0.23\linewidth}
         \centering
         \includegraphics[width=0.95\linewidth]{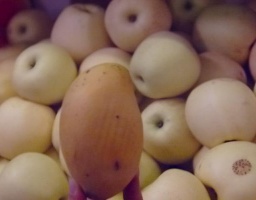}
     \end{subfigure}%
     \begin{subfigure}[b]{0.23\linewidth}
         \centering
         \includegraphics[width=0.95\linewidth]{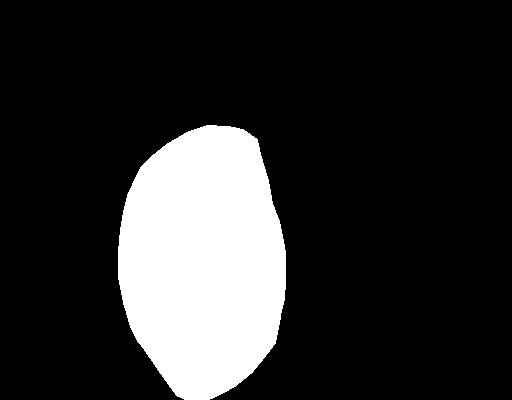}
     \end{subfigure}%
     \begin{subfigure}[b]{0.23\linewidth}
         \centering
         \includegraphics[width=0.95\linewidth]{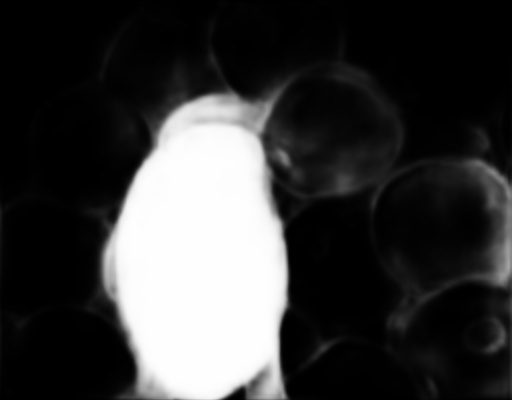}
     \end{subfigure}%
     \begin{subfigure}[b]{0.23\linewidth}
         \centering
         \includegraphics[width=0.95\linewidth]{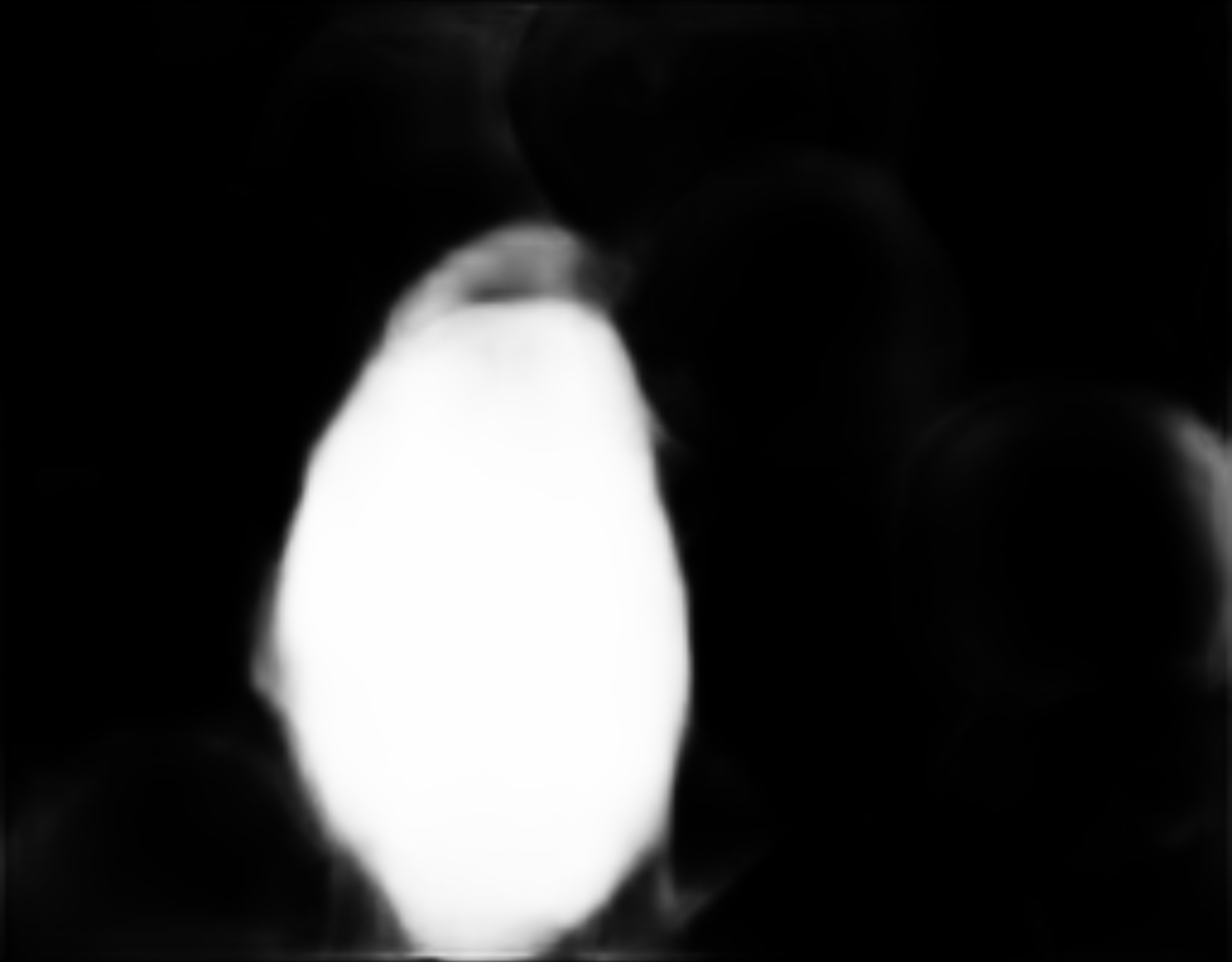}
     \end{subfigure}
     
     \begin{subfigure}[b]{0.23\linewidth}
         \centering
         \includegraphics[width=0.95\linewidth]{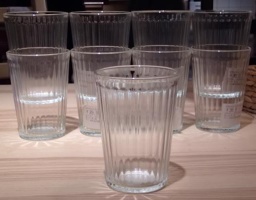}
     \end{subfigure}%
     \begin{subfigure}[b]{0.23\linewidth}
         \centering
         \includegraphics[width=0.95\linewidth]{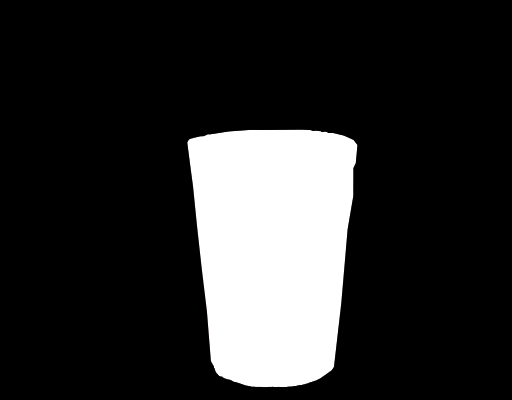}
     \end{subfigure}%
     \begin{subfigure}[b]{0.23\linewidth}
         \centering
         \includegraphics[width=0.95\linewidth]{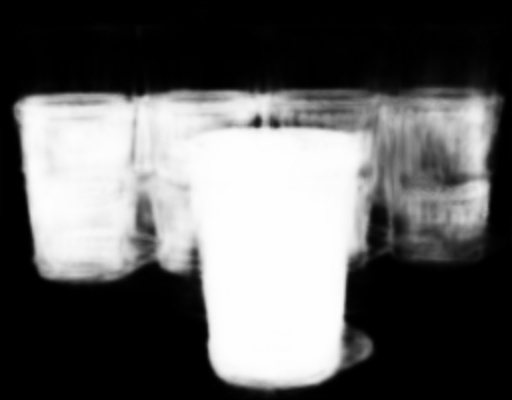}    
     \end{subfigure}%
     \begin{subfigure}[b]{0.23\linewidth}
         \centering
         \includegraphics[width=0.95\linewidth]{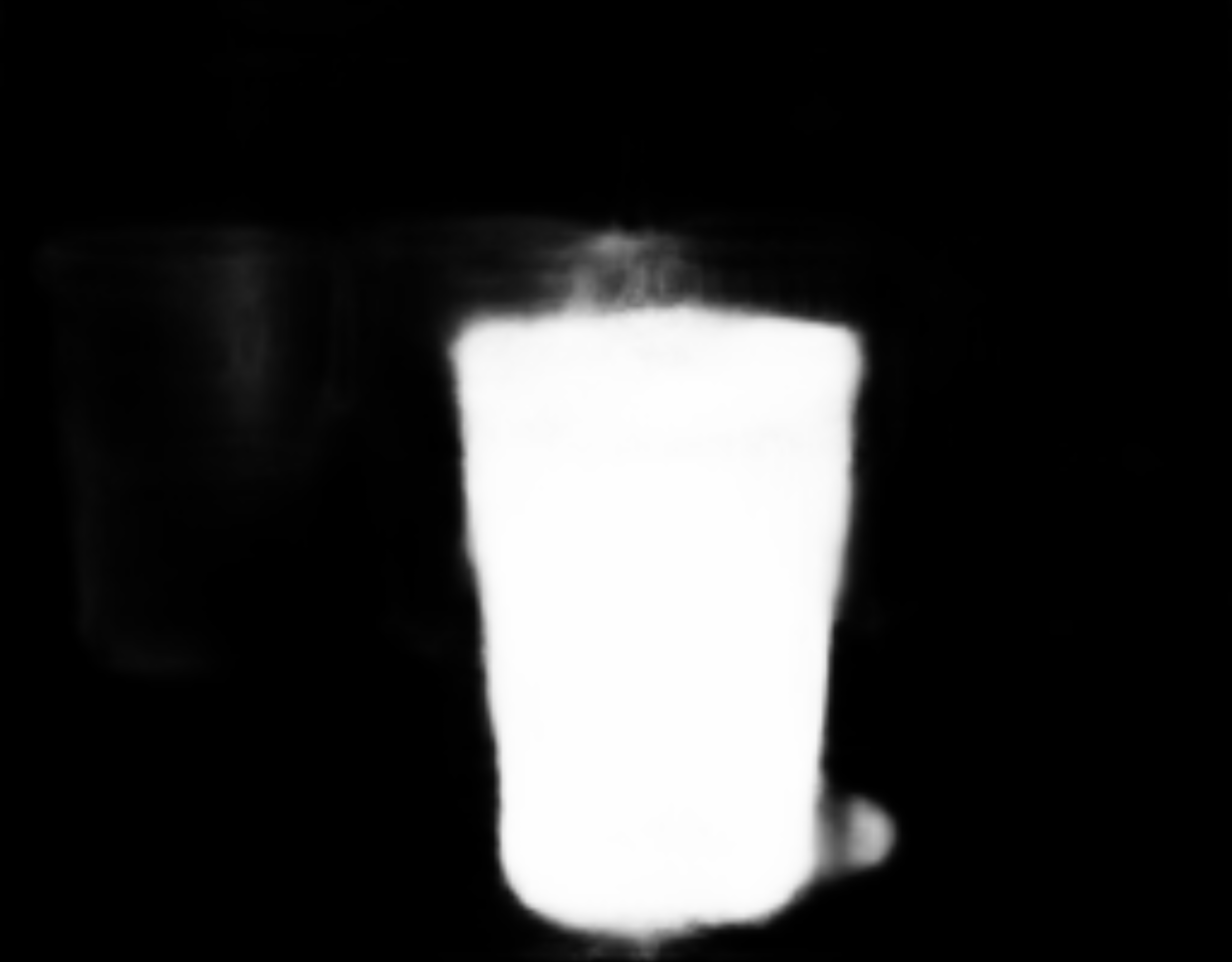}
     \end{subfigure}
     
     \begin{subfigure}[b]{0.23\linewidth}
         \centering
         \includegraphics[width=0.95\linewidth]{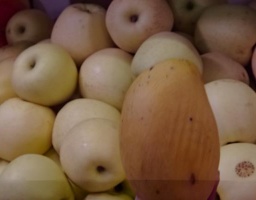}
     \end{subfigure}%
     \begin{subfigure}[b]{0.23\linewidth}
         \centering
         \includegraphics[width=0.95\linewidth]{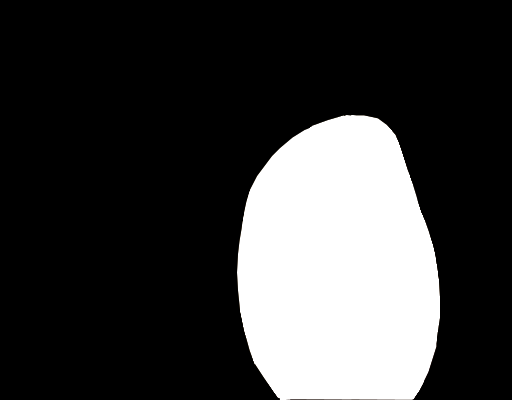}
     \end{subfigure}%
     \begin{subfigure}[b]{0.23\linewidth}
         \centering
         \includegraphics[width=0.95\linewidth]{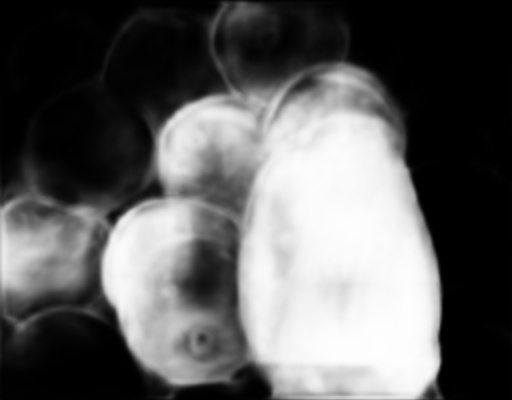}
     \end{subfigure}%
     \begin{subfigure}[b]{0.23\linewidth}
         \centering
         \includegraphics[width=0.95\linewidth]{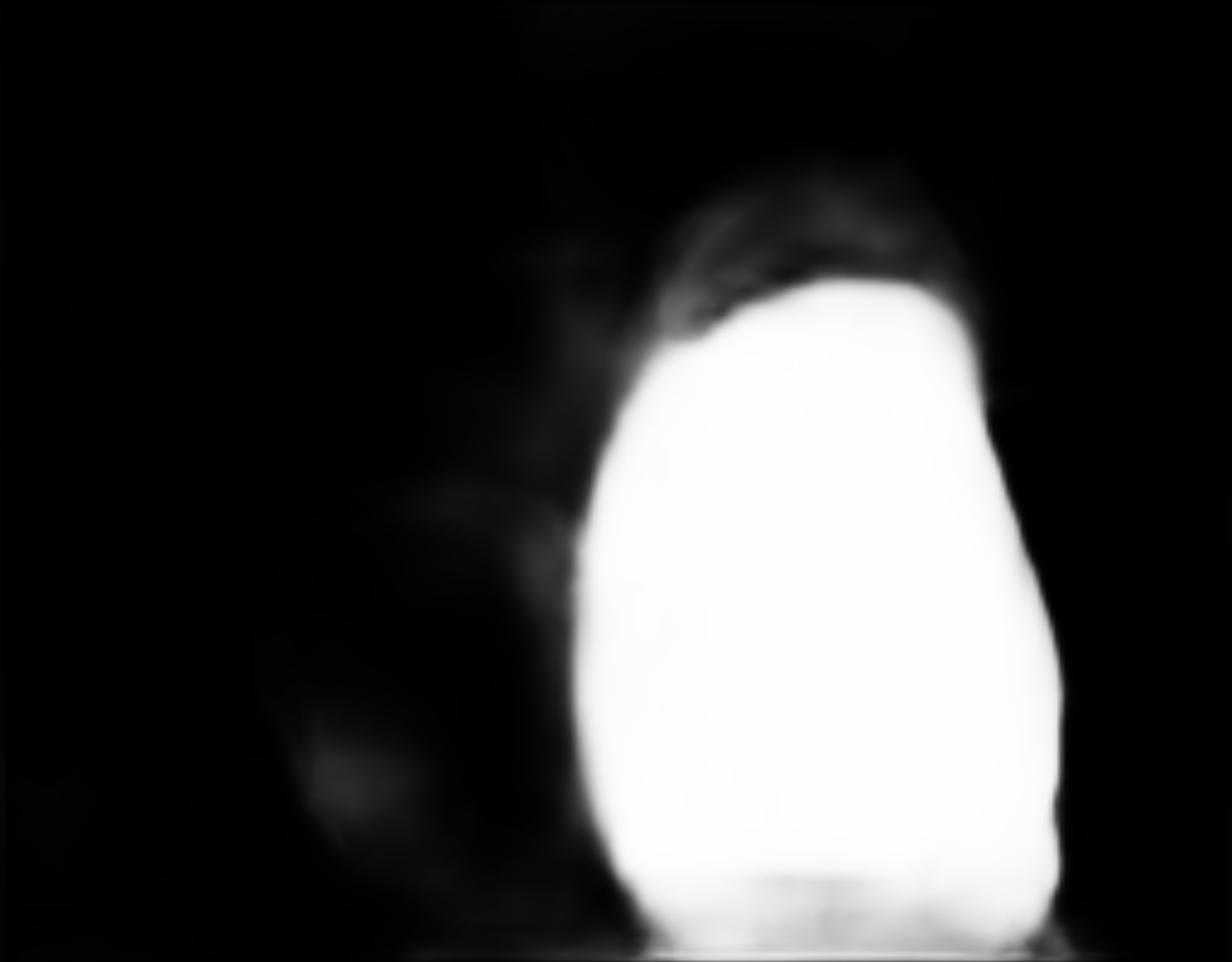}
     \end{subfigure}
     
     \begin{subfigure}[b]{0.23\linewidth}
         \centering
         \includegraphics[width=0.95\linewidth]{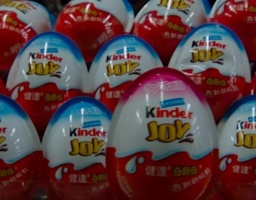}
     \end{subfigure}%
     \begin{subfigure}[b]{0.23\linewidth}
         \centering
         \includegraphics[width=0.95\linewidth]{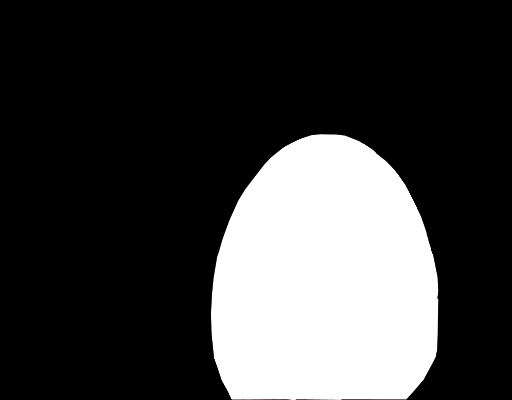}
     \end{subfigure}%
     \begin{subfigure}[b]{0.23\linewidth}
         \centering
         \includegraphics[width=0.95\linewidth]{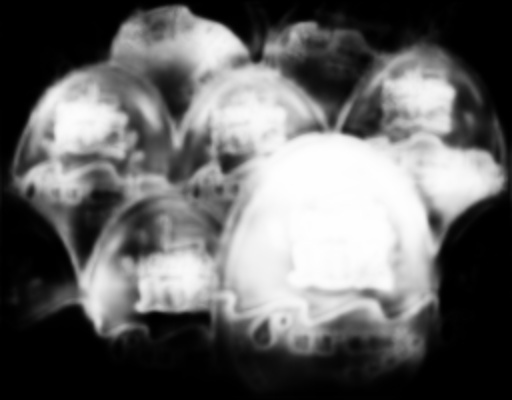}
     \end{subfigure}%
     \begin{subfigure}[b]{0.23\linewidth}
         \centering
         \includegraphics[width=0.95\linewidth]{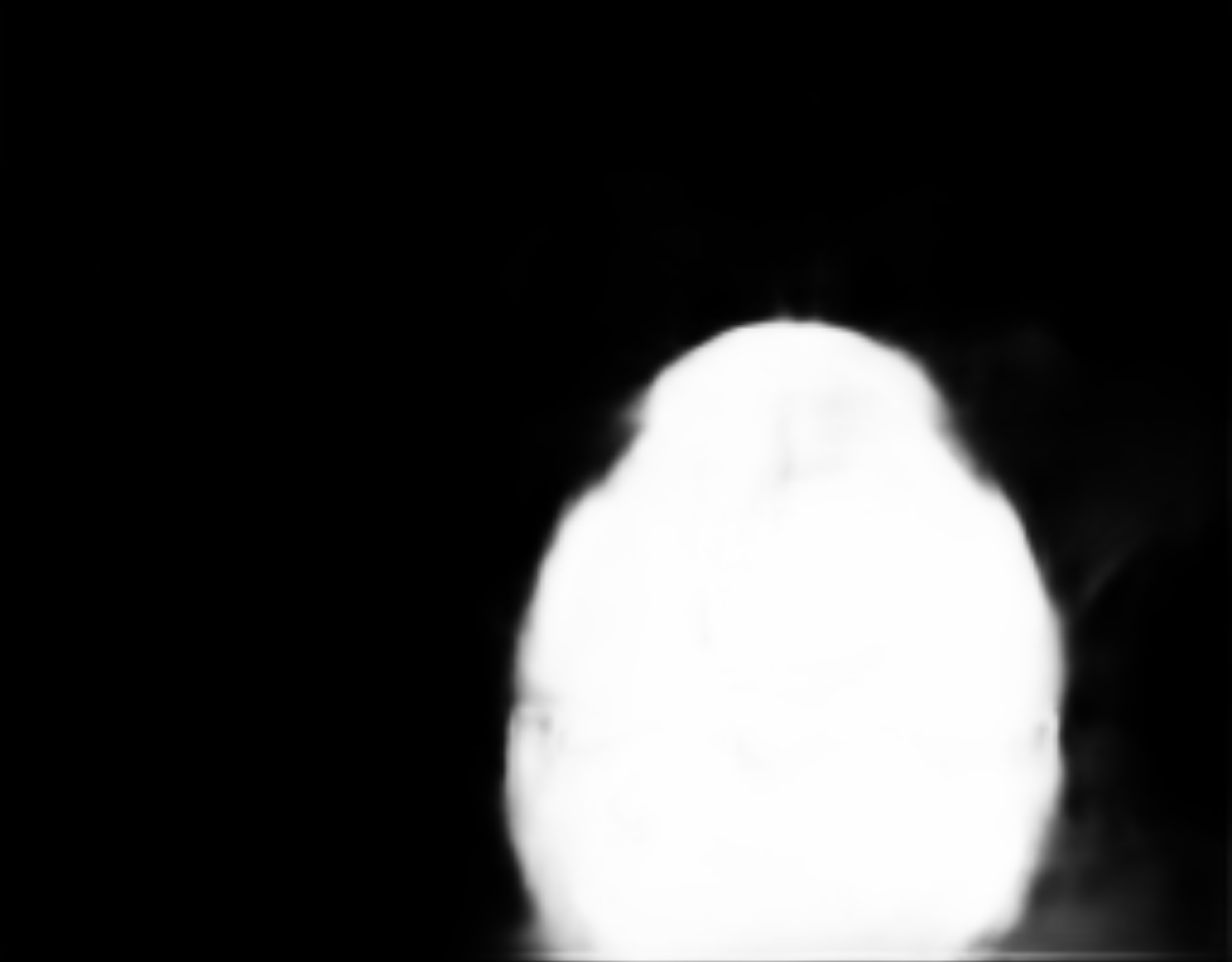}
     \end{subfigure}
     
     \begin{subfigure}[b]{0.23\linewidth}
         \centering
         \includegraphics[width=0.95\linewidth]{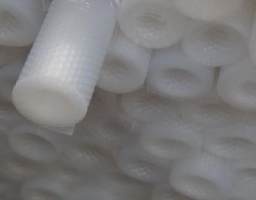}
     \end{subfigure}%
     \begin{subfigure}[b]{0.23\linewidth}
         \centering
         \includegraphics[width=0.95\linewidth]{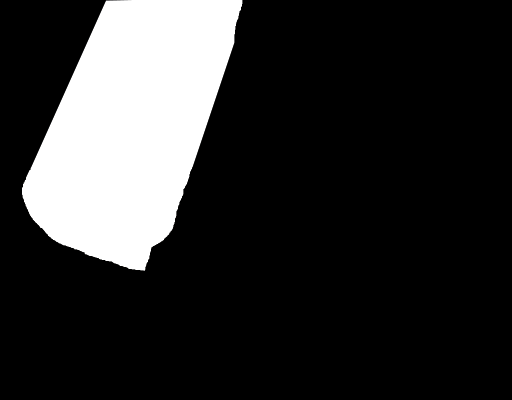}
     \end{subfigure}%
     \begin{subfigure}[b]{0.23\linewidth}
         \centering
         \includegraphics[width=0.95\linewidth]{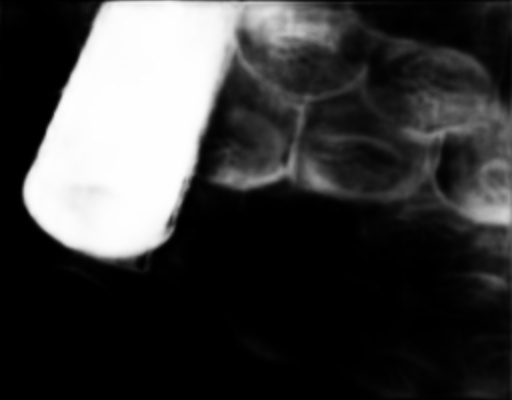}
     \end{subfigure}%
     \begin{subfigure}[b]{0.23\linewidth}
         \centering
         \includegraphics[width=0.95\linewidth]{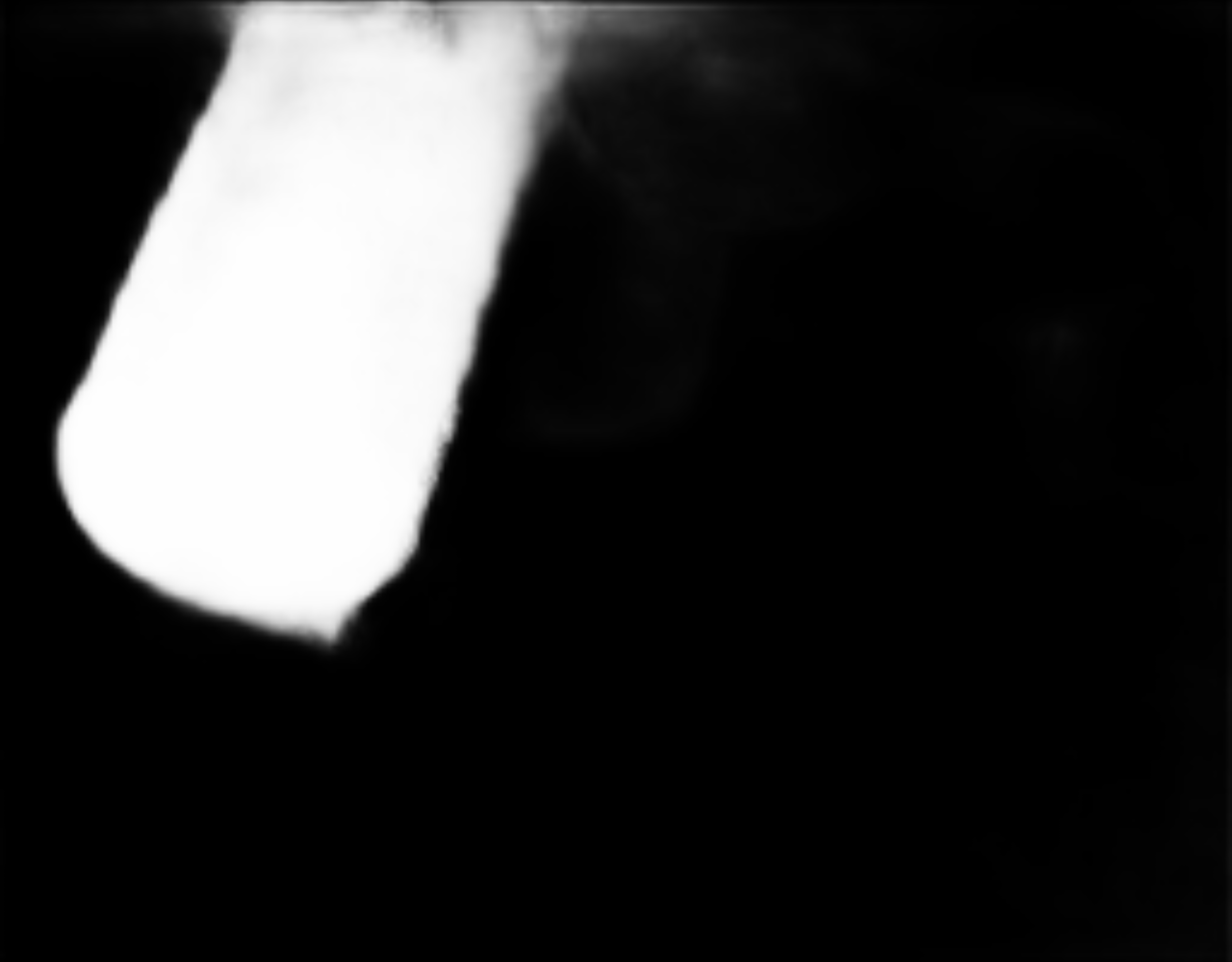}
     \end{subfigure}

     \begin{subfigure}[b]{0.23\linewidth}
         \centering
         \includegraphics[width=0.95\linewidth]{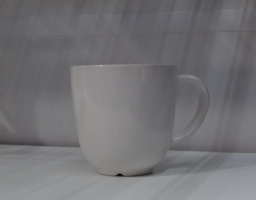}
         \caption{Center SAI}
         \label{sfi:sf1}
     \end{subfigure}%
     \begin{subfigure}[b]{0.23\linewidth}
         \centering
         \includegraphics[width=0.95\linewidth]{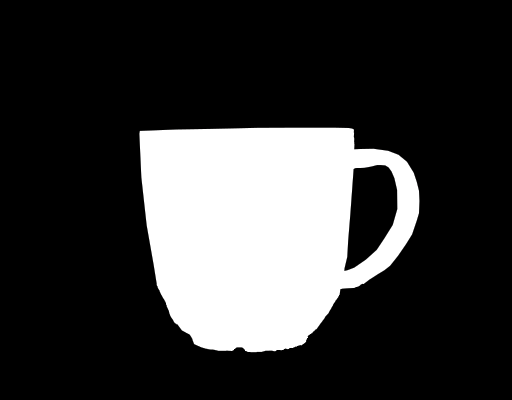}
         \caption{GT}
         \label{sfi:sf2}
     \end{subfigure}%
     \begin{subfigure}[b]{0.23\linewidth}
         \centering
         \includegraphics[width=0.95\linewidth]{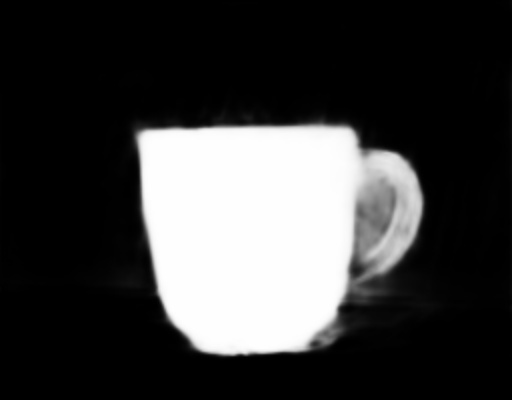}
         \caption{Base~\cite{zhao2019pyramid}}
         \label{sfi:sf3}
     \end{subfigure}%
     \begin{subfigure}[b]{0.23\linewidth}
         \centering
         \includegraphics[width=0.95\linewidth]{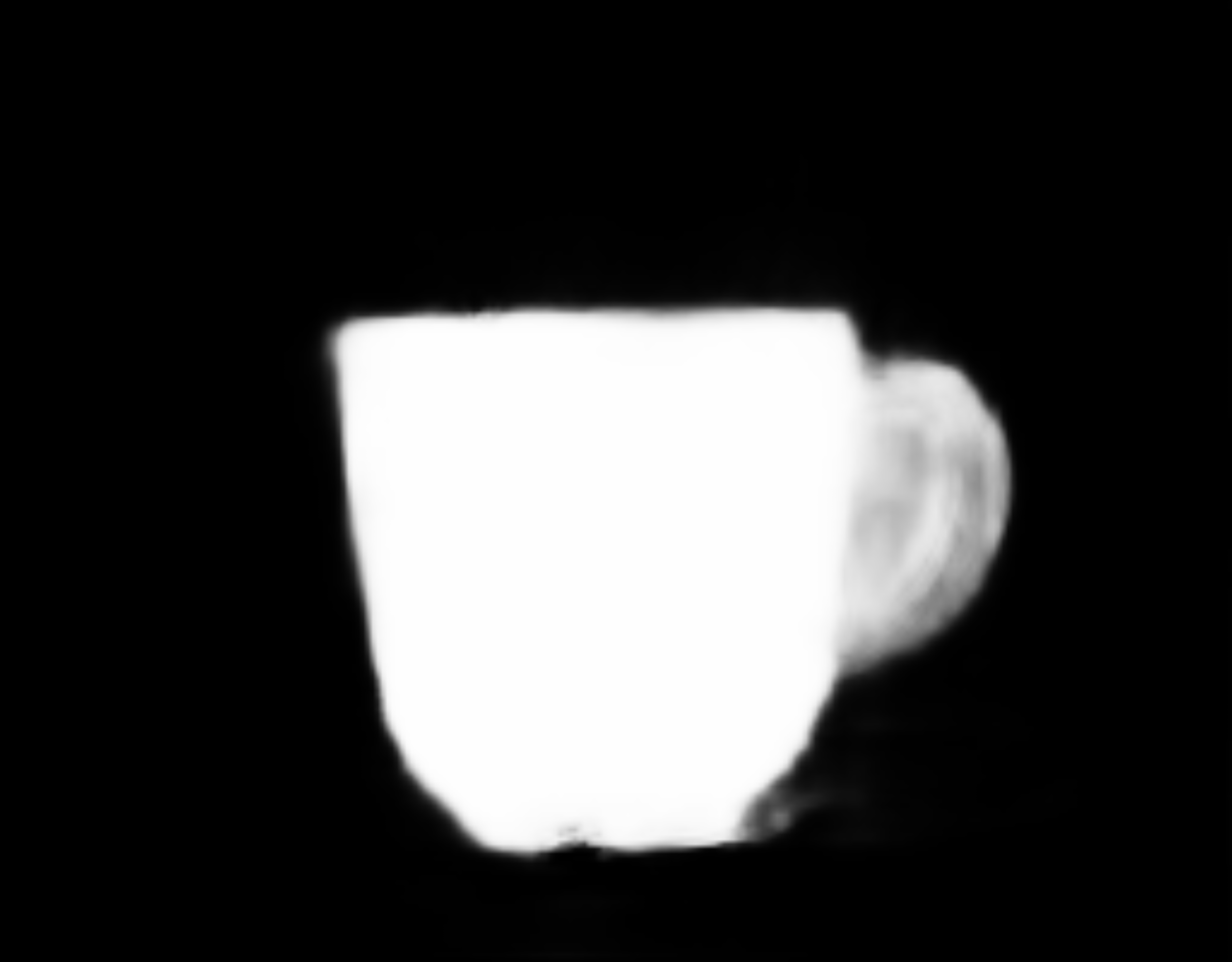}
         \caption{Ours}
         \label{sfi:sf4}
     \end{subfigure}     
    \caption{Comparison of saliency maps: (a) centre sub-aperture image (SAI) of the light field, (b) ground truth (GT), (c) base network, (d) our results in DUTLF-v2 dataset.As it can be seen in the results, our outputs are more closer to the ground truth compared to the base model\cite{zhao2019pyramid} outputs.}
    \label{fi:pictorial_results_dutlf}
    \vspace{-4ex}
\end{figure}

We employ the evaluation metrics $F_{\beta}$ measure (with $\beta^{2} = 0.3$ as suggested in \cite{achanta2009frequency}, MAE, and $F^{w}_{\beta}$ measure to compare the performance of the saliency detectors. The metrics $F_{\beta}$, $F^{w}_{\beta}$ and MAE are, respectively, defined as
\vspace{-1ex}
\begin{align}
    F_{\beta} &=\frac{(1+\beta^2)\times Precision \times Recall}{(\beta^2\times Precision) + Recall}, 
    \label{eq:fbeta} \\
    F_{\beta}^{w} &= \dfrac{(1+\beta^{2})\times Precision^{w} \times Recall^{w}}{\beta^{2}\times Precision^{w} + Recall^{w}}, \label{eq:wfbeta} \\
    MAE &=\dfrac{1}{W \times H} \sum_{i=1}^{W} \sum_{j=1}^{H} |P(i,j)-G(i,j)|,  \label{eq:MAE}
\end{align}
where $P(i,j)$ and $G(i,j)$ are the output saliency map of a saliency detector and the ground truth saliency map, respectively, and $w$ is an Euclidean distance based weighting function~\cite{ran_2014_how}. We present the performance achieved with the proposed, LFNet~\cite{zhang2020lightfield} and MTCN~\cite{zhang2020multi} light field saliency detectors and five other 2-D saliency detectors in LYTRO ILLUM \cite{zhang2020lightfield} dataset in Table \ref{ta:results}. Accordingly, performance of our saliency detector is superior compared to LFNet while is slightly behind compared to MTCN in terms of all the three metrics. We show the saliency maps of twelve light fields obtained with the proposed, LFNet \cite{zhang2020lightfield} and MTCN \cite{zhang2020multi} light field saliency detectors in Figure~\ref{fi:pictorial_results} for qualitative comparison. Our saliency maps are closer to the ground truth compared to those of LFNet and comparable with the salience maps achieved with MTCN. Furthermore, our method greatly outperforms the accuracy obtained with the base model, i.e., the 2-D saliency detector in~\cite{zhao2019pyramid}. We also present the performance of the proposed light field saliency detector achieved for the DUTLF-v2 \cite{DUTLF} dataset in comparison to state-of-the-art 2-D saliency detectors, DLFS~\cite{piao2019deep} and student networks proposed in~\cite{DUTLF} and light field saliency detectors, LFNet~\cite{zhang2020lightfield}, DisenFusion~\cite{disenfusion}, ATAFNet~\cite{ATAFNet}, CPFP~\cite{CFPP}, MOLF~\cite{MoLF} and teacher network proposed in~\cite{DUTLF} in Table \ref{ta:results_dutlf}. Our method achieves second place in $F_{\beta}$ measure and achieves comparable performance in other measures. More importantly, our model is much smaller in size compared to other models except student~\cite{DUTLF} network. Accordingly, our methods provides \emph{a significant reduction in memory requirement}, especially compared to state-of-the-art light field saliency detectors of all the three categories: focal stack, RGB-D and light field. Therefore, our model is appropriate to be implemented in for resource constrained devices. Furthermore, we present saliency maps of our model for four light fields in Figure \ref{fi:pictorial_results_dutlf}. Note that outputs of our model are more closer to the ground truth compared to those of the base model.
\begin{table*}[t!]
\caption{Comparison with state-of-the-art light field saliency detectors, LFNet \cite{zhang2020lightfield}
, MTCN \cite{zhang2020multi} and RGB saliency detectors, NLDF \cite{NDLF}, PAGE-Net\cite{PAGE}, GCPANet \cite{GCPA} and SODGAN \cite{SODGAN}. Our results surpass LFNet~\cite{zhang2020lightfield}, DLFS\cite{piao2019deep}, and are slightly behind MTCN~\cite{zhang2020multi}.}
\label{ta:results}
\setlength{\tabcolsep}{3pt}
\vspace{-3ex}
\begin{center}
\scriptsize

\begin{tabular}{|l| c c c c c c| c c c|} 
 \hline
 \multirow{2}{*}{ } &\multicolumn{6}{c|}{2-D} &\multicolumn{3}{c|}{Light field}\\
 Metric & NDLF\cite{NDLF} & PAGE-Net \cite{PAGE} & GCPA Net \cite{GCPA} & SCRNet \cite{SCRNet} & SODGAN \cite{SODGAN} & DLFS \cite{piao2019deep} & LFNet~\cite{zhang2020lightfield} & MTCN~\cite{zhang2020multi} & Ours \\  
 \hline
 $F_{\beta}$ & 0.7866 & 0.8047 & 0.8306 & 0.8473 & 0.8313 &0.7391& 0.8116 & 0.8729 & 0.8558  \\ 
 
 $F^{w}_{\beta}$ & 0.7299 & 0.7826 & 0.8100 & 0.8097 & 0.7969 & 0.6655 & 0.7540 & 0.8534 & 0.7671  \\

 MAE  & 0.0764 & 0.0723 & 0.0580 & 0.0551 & 0.0624& 0.0843 &0.0551 & 0.0483 & 0.0541  \\
 \hline
\end{tabular}
\end{center}
\vspace{-3ex}
\end{table*}
\normalsize

\begin{table*}[t!]
\caption{Comparison with state-of-the-art lightfield saliency detectors in DUTLF-v2\cite{DUTLF} dataset and our model has comparable performance and lags slightly behind Teacher model\cite{DUTLF} surpassing all the other algorithms in terms of $ F_{\beta}$ measure.}
\label{ta:results_dutlf}
\setlength{\tabcolsep}{3pt}
\vspace{-4ex}
\begin{center}
\scriptsize

\begin{tabular}{|c|c c| c c c| c c| c c|} 
 \hline
 \multirow{2}{*}{ } &\multicolumn{2}{c|}{2-D} &\multicolumn{3}{c|}{RGB-D} &\multicolumn{2}{c|}{Focal stack}&\multicolumn{2}{c|}{Light field}\\
 Metric & DLFS \cite{piao2019deep}& Student \cite{DUTLF} & DisenFusion\cite{disenfusion} & ATAFNet\cite{ATAFNet} & CPFP\cite{CFPP} & Teacher \cite{DUTLF} &  MoLF~\cite{MoLF} & LFNet \cite{zhang2020lightfield}   & Ours \\ 
 \hline
 $F_{\beta}$ &0.684 &0.813 & 0.686 & 0.808 & 0.707 & 0.852 & 0.723 &  0.803 &   0.8491 \\ 
 
 $F^{w}_{\beta}$ &0.641 & 0.771 & 0.636 & 0.775 & 0.629 & 0.792 & 0.709 & 0.786 &   0.7279 \\
 
 MAE   & 0.087 & 0.055 & 0.093 & 0.051 & 0.075 & 0.050 & 0.065 & 0.049 &   0.052  \\
  Size(M)   & 119 & 47 & 166 & 291.5  & 278 & 92.5 &  186.6 & 175.8 &  63  \\
 \hline
\end{tabular}
\end{center}
\vspace{-3ex}
\end{table*}
\normalsize

We present the computational time required by each light field saliency detector to process a light field in the LYTRO ILLUM dataset. Here, we consider only the saliency detectors LFNet~\cite{zhang2020lightfield} and MTCN~\cite{zhang2020multi}, which employ the full light field, for a fair comparison with the proposed saliency detection method. We present the computational time required by each light field saliency detector in Table~\ref{ta:speed} for both CPU and GPU implementations. Our saliency detector is \emph{$25$ times faster} than the LFNet in the CPU implementation, and \emph{require $55\%$ and $40\%$ less time} compared to LFNet and MTCN, respectively, for GPU implementation. Here, we present an approximated value for MTCN obtained based on the computational time reported in~\cite{zhang2020multi} ($1.2601$ s) for an implementation using a Nvidia Tesla P100 GPU. Accordingly, the proposed light field saliency detector \emph{significantly outperforms} state-of-the-art light field saliency detectors, and achieves \emph{near real-time} processing even in the CPU implementation. 

\begin{table}[t!]
\caption{Computational time required to process a light field:  our saliency detector is significantly faster than the state-of-the-art light field saliency detectors.}

\vspace{-3ex}
\label{ta:speed}

\begin{center}
 \begin{tabular}{|c c c|} 
 \hline
 Method  & i9-9900K & RTX-2080TI\\ [0.5ex] 
 \hline
 LFNet~\cite{zhang2020lightfield}  & 10.4813 s & 0.5321 s \\

 MTCN~\cite{zhang2020multi}  & - & 0.3989$^*$ s \\

 Ours  & 0.4175 s & 0.2381 s\\
 \hline
 \multicolumn{3}{l}{\footnotesize $^*$approximated}
\end{tabular}
\end{center}
\vspace{-5ex}
\end{table}

\subsection{Ablation Study}
\label{asec:Ablation}
\begin{figure}[H]
     \centering
     \begin{subfigure}[b]{0.33\linewidth}
         \centering
         \includegraphics[width=0.95\linewidth]{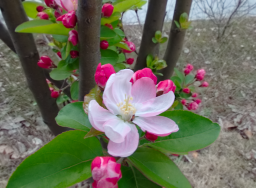}
     \end{subfigure}%
     \begin{subfigure}[b]{0.33\linewidth}
         \centering
         \includegraphics[width=0.95\linewidth]{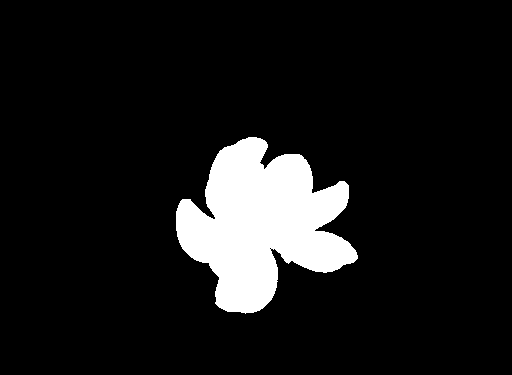}
     \end{subfigure}%
     \begin{subfigure}[b]{0.33\linewidth}
         \centering
         \includegraphics[width=0.95\linewidth]{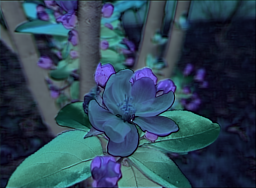}
     \end{subfigure}%
     
     \begin{subfigure}[b]{0.33\linewidth}
         \centering
         \includegraphics[width=0.95\linewidth]{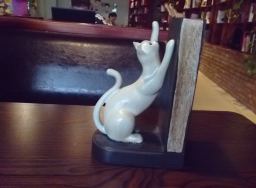}
     \end{subfigure}%
     \begin{subfigure}[b]{0.33\linewidth}
         \centering
         \includegraphics[width=0.95\linewidth]{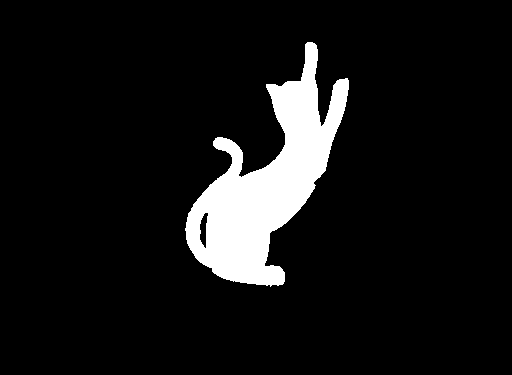}
     \end{subfigure}%
     \begin{subfigure}[b]{0.33\linewidth}
         \centering
         \includegraphics[width=0.95\linewidth]{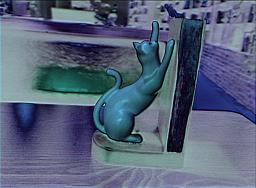}
     \end{subfigure}%

     \begin{subfigure}[b]{0.33\linewidth}
         \centering
         \includegraphics[width=0.95\linewidth]{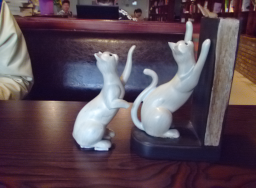}
     \end{subfigure}%
     \begin{subfigure}[b]{0.33\linewidth}
         \centering
         \includegraphics[width=0.95\linewidth]{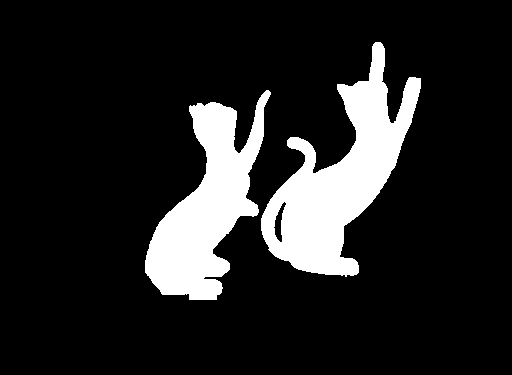}
     \end{subfigure}%
     \begin{subfigure}[b]{0.33\linewidth}
         \centering
         \includegraphics[width=0.95\linewidth]{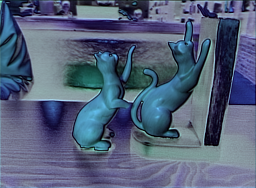}
     \end{subfigure}%
     
     \begin{subfigure}[b]{0.33\linewidth}
         \centering
         \includegraphics[width=0.95\linewidth]{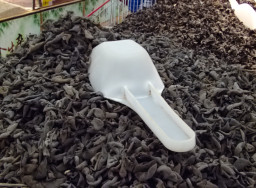}
     \end{subfigure}%
     \begin{subfigure}[b]{0.33\linewidth}
         \centering
         \includegraphics[width=0.95\linewidth]{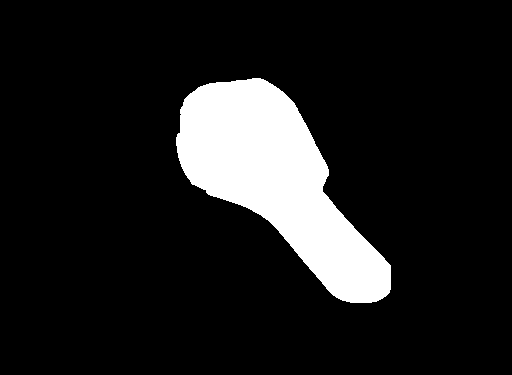}
     \end{subfigure}%
     \begin{subfigure}[b]{0.33\linewidth}
         \centering
         \includegraphics[width=0.95\linewidth]{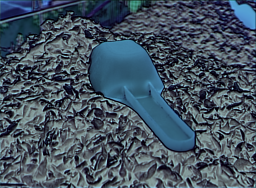}
     \end{subfigure}%
     
     \begin{subfigure}[b]{0.33\linewidth}
         \centering
         \includegraphics[width=0.95\linewidth]{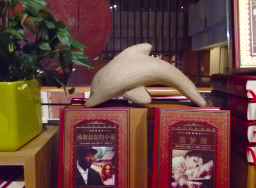}
         \caption{Center SAI}
     \end{subfigure}%
     \begin{subfigure}[b]{0.33\linewidth}
         \centering
         \includegraphics[width=0.95\linewidth]{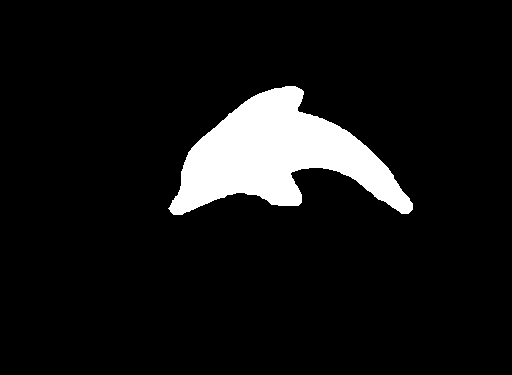}
         \caption{GT}
     \end{subfigure}%
     \begin{subfigure}[b]{0.33\linewidth}
         \centering
         \includegraphics[width=0.95\linewidth]{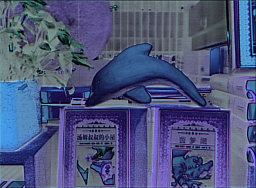}
         \caption{FEE}
     \end{subfigure}%
    \caption{This figure contains (a) center sub aperture image, (b) ground truth and (c) scaled outputs from the FEE module. We observe that the regions with the higher disparity are emphasized from the background by borders. This verifies that the FEE module successfully encodes the features required for for salient object detection.}
    \label{fi:fee}
\end{figure}

In this section, we present the comparison between the base model~\cite{zhao2019pyramid} and our architecture with the FEE module. In Table \ref{ta:ablation}, we compare the performance increment obtained through the FEE block in LYTRO ILLUM~\cite{zhang2020lightfield} and DUTLF~\cite{DUTLF} datasets. As we can see from the results, FEE model provides a \emph{significant boost} in performance compared to the base model~\cite{zhao2019pyramid} in both cases. It is worthwhile to note that the $F_{\beta}$ measure of our method could be further improved by employing the original base model with MAE$=0.0405$~\cite{zhao2019pyramid} than our trained model with MAE$=0.0698$. In Figure \ref{fi:fee}, we present the scaled outputs from the FEE block for light fields of the LYTRO ILLUM dataset~\cite{zhang2020lightfield}. We can observe in the outputs of the FEE module that the regions with the higher disparity, i.e., salient objects in the foreground, are emphasized from the background by a border. This verifies that FEE module successfully encodes features of a light field required for salient object detection. This feature embedding helps to accurately segment the salient regions while avoiding computationally heavy backbones.

\begin{table}
\caption{Comparison of $F_{\beta}$ scores between base-model and the neural network.As depicted in the table, addition of FEE module gives a huge boost to the performance of the model for both datasets}
\vspace{-3ex}
\label{ta:ablation}

\begin{center}
 \begin{tabular}{|c c c|} 
 \hline
 Method & LYTRO ILLUM & DUTLF-V2 \\ [0.5ex] 
 \hline
 Base model & 0.7322 & 0.7275  \\ 
 
 FEE $+$ Base model & 0.8558 & 0.8491  \\ 
 \hline
\end{tabular}
\end{center}
\vspace{-2ex}
\end{table}

\section{Conclusion and Future Work}
\label{sec:cnfw}

We proposed a fast and accurate light field saliency detector that feeds carefully computed light field features to a saliency detector with an attention mechanism. It is fast and runs on an i9 CPU at approximately $2$ light fields/s and on a 2080Ti GPU at 4 light fields/s leading to near real-time processing. Furthermore, the memory requirement of our model is significantly lower compared to state-of-the-art light field saliency detectors making our model appropriate for resource constrained devices. The accuracy of our model surpasses most of the existing methods, and is only slightly inferior to a very recent work. The speed is due to faster feature extraction that constrains light field processing only to the FEE module and using a single stream without resorting to recurrent networks. The high accuracy is due to the light field saliency specific feature extractor and the use of an attention mechanism. Our work brings light field saliency detection closer to real-time implementations which would enable, e.g., cameras to refocus on objects of interest.

Future directions include making the network faster and more accurate by changing or improving the 2-D detector backbone and FEE module. Adapting this method to other computer vision tasks which benefit from the angular information embedded in light fields and lack reasonably-sized datasets---such as, material recognition, segmentation, and object detection---which use 2-D-input neural networks would also be interesting.

\bibliography{mybibfile}

\end{document}